\documentclass[10pt,twocolumn,letterpaper]{article}

\usepackage{iccv}
\usepackage{times}
\usepackage{epsfig}
\usepackage{graphicx}
\usepackage{amsmath}
\usepackage{amssymb}

\usepackage{algorithm}
\usepackage{algorithmicx}
\usepackage{algpseudocode}
\usepackage{subfigure}
\usepackage{multirow}

\usepackage{diagbox}
\usepackage{amsthm}
\usepackage{bm}
\renewcommand{\mathbf}{\boldsymbol}

\def\x{\mathbf{x}}

\def\p{\mathbf{p}}

\def\y{\mathbf{y}}

\def\f{\mathbf{f}}

\def\p{\mathbf{p}}

\DeclareMathOperator*{\argmax}{arg\,max}
\DeclareMathOperator*{\argmin}{arg\,min}

\usepackage{dsfont}

\usepackage[pagebackref=true,breaklinks=true,letterpaper=true,colorlinks,bookmarks=false]{hyperref}

\iccvfinalcopy 

\def\iccvPaperID{3002} 

\begin{document}
\newcolumntype{L}[1]{>{\raggedright\arraybackslash}p{#1}}
\newcolumntype{C}[1]{>{\centering\arraybackslash}p{#1}}
\newcolumntype{R}[1]{>{\raggedleft\arraybackslash}p{#1}}
\title{Gradual Domain Adaptation via Self-Training of Auxiliary Models}

\author{Yabin Zhang$^1$ \quad Bin Deng$^2$ \quad Kui Jia$^2$ \quad Lei Zhang$^1$\\
$^1$Hong Kong Polytechnic University \quad $^2$South China University of Technology \\
{\tt\small \{csybzhang,cslzhang\}@comp.polyu.edu.hk \quad bindeng.scut@gmail.com \quad kuijia@scut.edu.cn  } 
}

\maketitle
\ificcvfinal\thispagestyle{empty}\fi

\begin{abstract}
	Domain adaptation becomes more challenging with increasing gaps between source and target domains. Motivated from an empirical analysis on the reliability of labeled source data for the use of distancing target domains, we propose self-training of auxiliary models (AuxSelfTrain) that learns models for intermediate domains and gradually combats the distancing shifts across domains. We introduce evolving intermediate domains as combinations of decreasing proportion of source data and increasing proportion of target data, which are sampled to minimize the domain distance between consecutive domains. 
	Then the source model could be gradually adapted for the use in the target domain by self-training of auxiliary models on evolving intermediate domains. We also introduce an enhanced indicator for sample selection via implicit ensemble and extend the proposed method to semi-supervised domain adaptation. Experiments on benchmark datasets of unsupervised and semi-supervised domain adaptation verify its efficacy.
\end{abstract}

\section{Introduction}  \label{SecIntro}
Deep neural network (DNN) has achieved great success on various tasks. However, it requires large amounts of labeled training data and generalizes poorly to new data domains. If we directly apply the trained DNN to a new data domain, degenerated performance is probably encountered. At the same time, it is expensive to collect labeled training data for the new domain. Therefore, there is a strong motivation to transfer knowledge from the labeled source domain to the unlabeled target one, which falls in the realm of transfer learning \cite{transfer_survey} or domain adaptation (DA) \cite{dan,dann}.

Seminal DA theories \cite{ben2007analysis,ben2010theory,zhang2019bridging} suggest that the target error could be minimized by reducing the source error and the distribution discrepancy across domains. Inspired by these theoretical results, DA methods are typically proposed to minimize the domain discrepancy by learning domain invariant representations \cite{dan,dann,cada,mcd,symnets}.  For example,  measurable distance metrics are calculated and minimized across domains \cite{dan,deep_coral}. However, the DA problems are getting more difficult as domain gaps between source and target domains enlarge, which brings new challenges to traditional DA methods \cite{bobu2018adapting}.

We empirically investigate the reliability of labeled source data for the use of distancing target domains in Section \ref{Sec:motivation}. As the distance between the source and target domains enlarges, the model trained on labeled source data gives the decreasing performance on the unlabeled target data, as showed in Figure \ref{Fig:score_acc_dis_vs_shift}. This indicates that DA could be facilitated and may have improved results if some reliable auxiliary models for intermediate domains are available.

Fortunately, as theoretically justified in \cite{kumar2020understanding}, given two domains with small distribution divergence and a reliable model on one domain, the reliable model for the other one could be achieved by self-training. 
In other words, reliable auxiliary models could be achieved via self-training if intermediate domains, where the divergence between consecutive domains is small, are available. 
In the literature \cite{xie2015transfer,bobu2018adapting,kumar2020understanding}, such expected intermediate domains are assumed to be readily available. However, the evolving intermediate domains may be not given in practice, such as in standard benchmark datasets \cite{home,peng2019moment}. 

To this end, we introduce intermediate domains as combinations of certain proportions of source data and  target data, inspired by \cite{chopra2013dlid}. Specifically, starting from source data and a well trained source model, we introduce an intermediate domain with small domain divergence to the source one by combining the majority of source data and a few number of target data. A reliable auxiliary model for the introduced intermediate domain could be achieved via self-training \cite{kumar2020understanding}. Given the previous intermediate domain and its reliable model achieved via self-training, we alternate between introducing another intermediate domain with small domain divergence to the previous one by combining less number of source data and more number of target data and getting the auxiliary model for the new intermediate domain via self-training. We finally end with the target data and gradually adapt the source model for it.

Considering that there may be an intra-domain divergence in one dataset \cite{mancini2018boosting,chen2019blending}, thus selecting which part of target and source data to construct intermediate domains matters. 
Given the starting point of labeled source data, target samples with smaller domain shift to source data should be selected first to minimize the domain divergence between consecutive domains; similarly, source samples with larger domain shift to target data should be dropped first, considering that target domain is the terminal of intermediate domains. We achieve this goal based on an empirical observation of the correlation between the domain shift and maximum prediction probability.  As illustrated in Figure \ref{Fig:score_acc_dis_vs_shift}, based on a model trained on labeled source data, the samples with smaller domain distance to the labeled one tend to present higher prediction probability. Therefore, we could construct the expected intermediate domain by selecting target samples with higher prediction probability based on the current auxiliary model, and dropping the source samples with lower prediction probability based on an introduced target prototype classifier. 
This sampling strategy significantly improves over the previous one \cite{chopra2013dlid}, where samples are randomly selected, as showed in Section \ref{Sec:experiment}.

Considering that the maximum prediction probability may be not reliable since DNNs may produce high probability for inputs far away from training data \cite{nguyen2015deep} , we propose an implicit ensemble scheme to improve the quality of predictive uncertainty to domain shift \cite{lakshminarayanan2016simple}.
As combinations of source and target data are adopted as intermediate domains in our AuxSelfTrain, we could easily extend the AuxSelfTrain to semi-supervised domain adaptation (SSDA) tasks \cite{ssda_mme}, where a few labeled target data are available, by treating the combination of source data and labeled target data as a pre-given intermediate domain and beginning our method from it. Empirical results on DA and SSDA benchmark datasets justify the efficacy of our AuxSelfTrain. 
We summarize our contributions as follows:
\begin{itemize}
	\item Motivated from an empirical analysis on the reliability of labeled source data for the use of distancing target domains, we propose for DA the  self-training of auxiliary models (AuxSelfTrain) that learns auxiliary models for intermediate domains and gradually combats the distancing shifts across domains.
	\item To technically implement AuxSelfTrain, we propose simple yet efficient sample selection strategies to select decreasing proportion of source data and increasing proportion of target data to construct evolving intermediate domains. 
	We also introduce an enhanced selection indicator by improving the quality of predictive uncertainty to domain shift via implicit ensemble.
	\item We conduct careful analyses on AuxSelfTrain and extend it to SSDA task. Empirical results on standard benchmark datasets of DA and SSDA justify the efficacy of our method.
\end{itemize}

\section{Related Works}
\noindent \textbf{Representative methods for domain adaptation}
The seminal DA theories \cite{ben2007analysis,ben2010theory,zhang2019bridging} suggest that we could minimize the target error by reducing the source error and the cross-domain discrepancy. Motivated by them, the distance minimization-based methods \cite{dan,deep_coral} and adversarial training-based ones \cite{dann,cada,mcd,symnets} have been proposed to minimize the domain discrepancy.
In the former one, the domain discrepancy is measured and minimized by measurable distance metrics, such as the maximum mean discrepancy \cite{dan} and correlation distance \cite{deep_coral}.  The latter one minimizes the domain discrepancy by playing minimax optimization \cite{gan} between the feature extractor and domain classifiers \cite{dann,cada} or task classifiers \cite{mcd,symnets}. 
However, as the domain gap enlarges, the performance of seminal DA methods degrades significantly \cite{bobu2018adapting}; additionally, there are  potential risks in the discrepancy minimization process \cite{chen2019transferability,zhao2019learning}. In contrary, we tackle DA by gradually adapt the source model for the use of target domain via self-training of auxiliary models on evolving intermediate domains.

\noindent \textbf{Domain adaptation with intermediate domains}
To facilitate the adaptation across source and distancing target domains, researches typically utilize intermediate domains as a bridge.
In the literature \cite{xie2015transfer,bobu2018adapting,kumar2020understanding}, intermediate domains are assumed to be readily available for the adaptation use. Although promising success has been achieved, the assumption of readily available intermediate domains may be not met and constrains its practical applications. Some methods \cite{gong2019dlow,cui2020gradually} try to introduce intermediate domains with generation models, and the generated intermediate samples are typically used as an additional regularization, rather than supporting auxiliary models as in our AuxSelfTrain. The work of \cite{chopra2013dlid} is closest to our method, but there are two crucial differences. Chopra \emph{et al.} \cite{chopra2013dlid} randomly sample target and source data to construct intermediate domains and the corresponding models are introduced with unsupervised learning, while we propose simple yet efficient sampling strategies for both domains and achieve auxiliary models via self-training, as suggested by \cite{kumar2020understanding}.

\noindent \textbf{Self-training} 
As a basic technique, self-training is initially introduced for semi-supervised learning (SSL) \cite{lee2013pseudo}, and then widely adopted in unsupervised learning \cite{caron2018deep}, SSL \cite{sohn2020fixmatch}, and DA \cite{zou2018unsupervised,Mei2020instance,kim2019self}. DA methods \cite{zou2018unsupervised,Mei2020instance,kim2019self} typically adopt self-training to adapt source model to target domain directly, while we, together with \cite{kumar2020understanding}, emphasize evolving intermediate domains for the success of self-training on DA.


\noindent \textbf{Semi-supervised domain adaptation} As a task with practical value, the SSDA is investigated with regularization \cite{daume2010frustratingly}, subspace learning \cite{yao2015semi} and smooth constraint \cite{donahue2013semi}. Recently, state-of-the-art results are achieved in \cite{ssda_mme, kim2020attract} by conducting adversarial entropy minimization with deep networks. We conduct experiments following \cite{ssda_mme} to justify the extension of AuxSelfTrain to SSDA.

\section{The Motivation: Analysis of the Source Reliability for Domain Adaptation} \label{Sec:motivation}

\begin{figure}	
	\centering
	\includegraphics[width=0.40\textwidth]{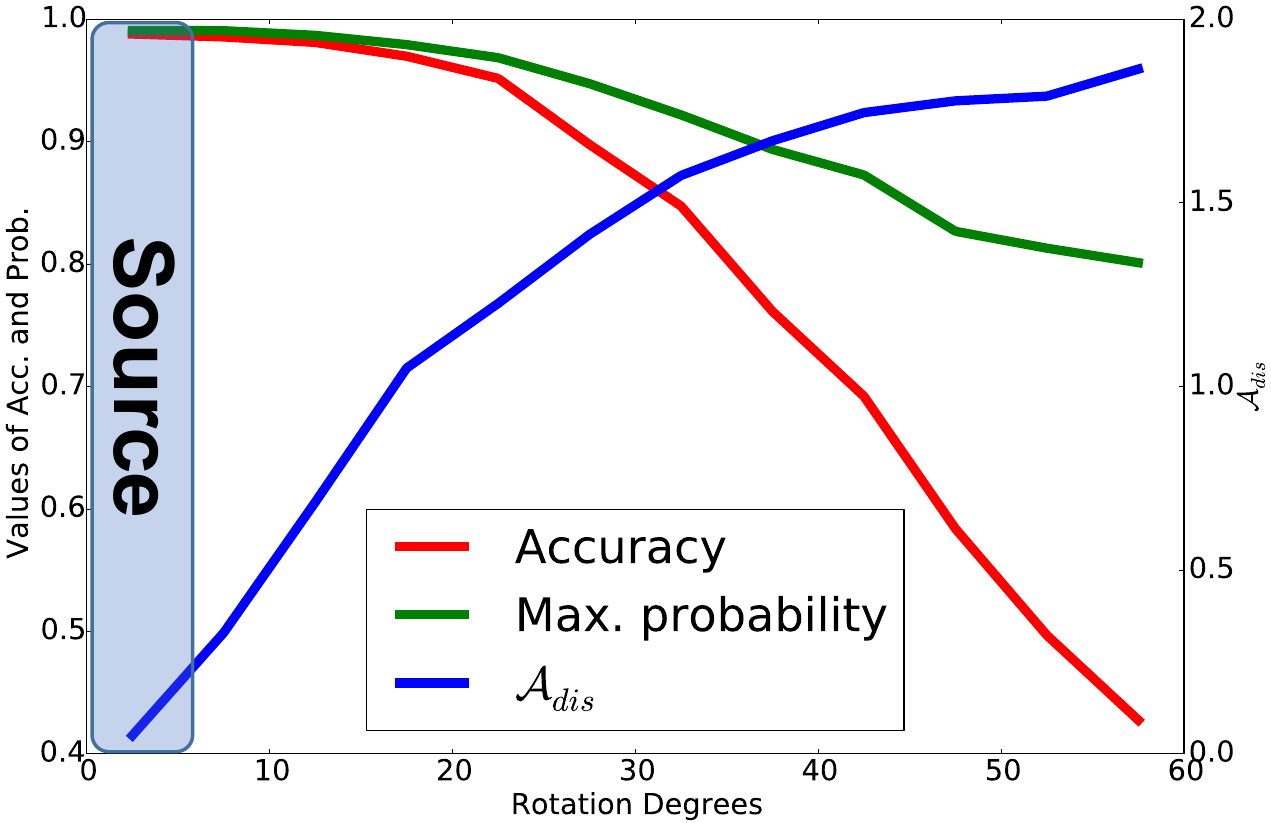}
	\caption{  An illustration of the reliability of the source data for the use of distancing target domains. As rotation degrees of target data get larger, i.e., the distribution divergence ($\mathcal{A}_{dis}$) across domains gets larger, the model trained on labeled source data presents decreasing performance on target data; at the same time, maximum prediction probability of target data gets smaller; this provides useful clues for the domain variation although the prediction is over-confident.    }   \label{Fig:score_acc_dis_vs_shift}
\end{figure}

We analyze the reliability of the source data for the use of distancing target domains on the Rotating MNIST dataset, which is a semi-synthetic dataset from the MNIST \cite{mnist}.  In Rotating MNIST, we strictly control the rotation degree of each sample in MNIST, where the domain shift is thus expected to be controlled. We randomly split the $50,000$ training samples of MNIST into the source and target domains of equal size. Then we randomly rotate the source samples by an angle from $0$ degree to $5$ degree to construct the labeled source domain. As for samples in the target domain, we randomly rotate them by an angle from $5r$ degree to $5r+5$ degree ($0 \leq r \leq 11$) to introduce various target domains. The enlarged distribution divergence between the source and target domains is expected by increasing the rotation degrees of target data, i.e., with larger $r$, which is also verified by the $\mathcal{A}_{dis}$ \cite{ben2007analysis} in Figure \ref{Fig:score_acc_dis_vs_shift}.  

As illustrated in Figure \ref{Fig:score_acc_dis_vs_shift}, the model trained on the labeled source data presents decreasing performance on unlabeled target data as rotation degrees of target data get larger, i.e., the domain divergence gets larger.
This also indicates that when some reliable auxiliary models for intermediate domains are available, domain adaptation could be facilitated and may have improved performance. 

Fortunately, the reliable auxiliary models for intermediate domains could be achieved through self-training, as theoretically proofed in \cite{kumar2020understanding}. Specifically, given an initial model with the low loss on the labeled training data and unlabeled data of a new domain with small domain distance to the labeled one, we could get a reliable model on the new domain via self-training (cf. Theorem 3.2 in \cite{kumar2020understanding}).
Thus, if intermediate datasets with gradually shifted distributions are available, we could gradually adapt the source model to the target domain along this path. 

However, intermediate datasets with evolving domain shift may be not available in practice, which prevents the application of such a promising strategy.
For example, the well ordered intermediate domains are not provided in standard DA benchmark datasets \cite{home,peng2019moment}. 
To this end, we make an empirical analysis and find that, based on a well-trained model, the maximum prediction probability (\ref{Equ:_target_score}) of unlabeled data gets smaller as their domain distance to the labeled training data gets larger, as illustrated in Figure \ref{Fig:score_acc_dis_vs_shift}; this provides useful clues for evolving intermediate domains and exactly motivates our method to be presented shortly. 

\begin{figure*}
	\centering
	\includegraphics[width=0.85\textwidth]{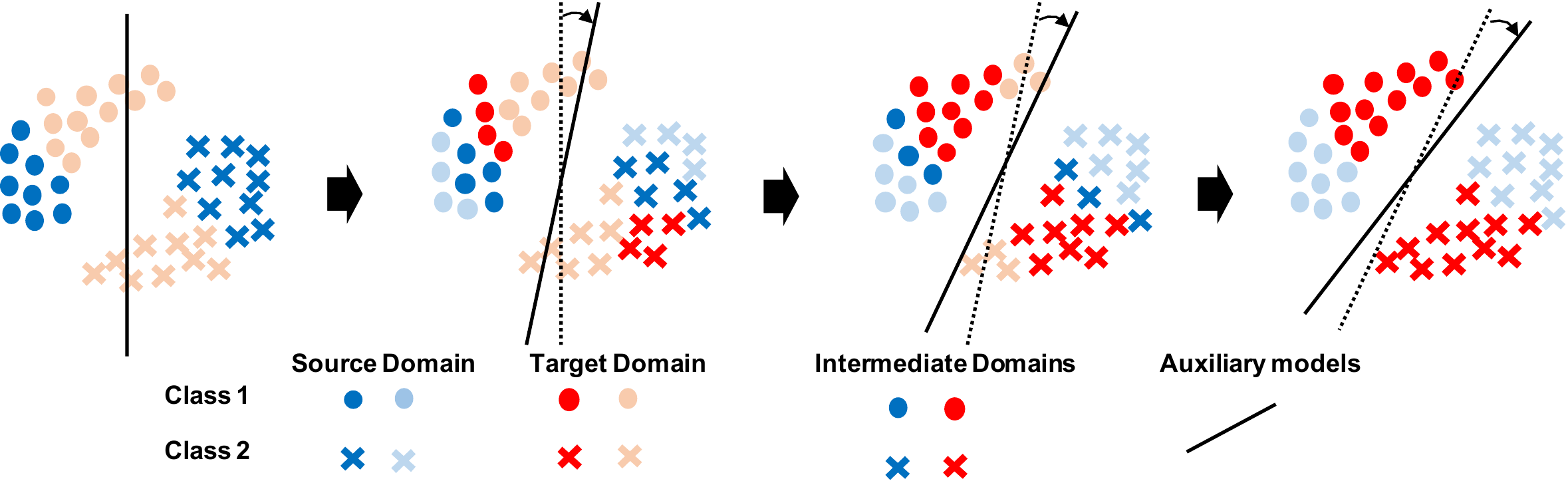}
	\caption{The framework of our proposed AuxSelfTrain that learns auxiliary models for intermediate domains and gradually adapts the source model for the use in the target domain. Best viewed in color.  }  \label{Fig:framework}
\end{figure*}

\section{The Proposed Method} \label{Sec:method}

In DA tasks, we have $n_s$ labeled source data $\mathcal{S} = \{ \x_i^s, y_i^s \}_{i=1}^{n_s}$ randomly sampled from a source domain $\mathcal{P}$ and $n_t$ unlabeled target data $\mathcal{T} = \{ \x_i^t \}_{i=1}^{n_t}$ randomly sampled from a related but different target domain $\mathcal{Q}$, where $y_i^s \in  \{1, \cdots, K \} $ and $K$ is the number of categories. 
The goal of DA is to minimize the target risk $\mathbb{E}_{(\x^t, y^t) \sim \mathcal{Q}} \mathcal{L}(\f(\x^t), y^t)$ for a certain loss $\mathcal{L}$ by learning a model $\f: \mathcal{X} \to [0,1]^K$ that maps any input instance $\x \in \mathcal{X}$ to the category probabilities.

\noindent\textbf{Self-training} is a popular method in both DA \cite{zou2018unsupervised, kumar2020understanding} and SSL \cite{lee2013pseudo,sohn2020fixmatch}. Take the DA task as an example, the self-trained method starts with a model $\f_s$ trained on the labeled source data:
\begin{equation} \label{Equ:source_pre_training}
\f_s = \argmin_{\f \in \mathcal{F}} \mathbb{E}_{\{ \x^s, y^s \} \in \mathcal{S}} \mathcal{L} (\f(\x^s), y^s),
\end{equation}
where $\mathcal{F}$ is the space of the model and $\mathcal{L} (\f(\x), y) = - \log f_y(\x)$ is the cross-entropy loss; $f_k(\x)$ denotes the $k$-th entry of $\f(\x)$.

Given the pre-trained model $\f_s$ and unlabeled data $\mathcal{T}$, self-training assigns a pseudo label to each sample $\x^t \in \mathcal{T}$ as $o(\f_s(\x^t)) = \argmax_{k} f_{s,k}(\x^t)$.
Then the target model $\f_t$ could be achieved with this pseudolabeled dataset $\{ \x_i^t, o(\f_s(\x^t_i)  \}_{i=1}^{n_t}$ as:
\begin{equation} \label{Equ:frame_da_ST}
\f_t = \argmin_{\f \in \mathcal{F}} \mathbb{E}_{\x^t \in \mathcal{T}} \mathcal{L}(\f(\x^t), o(\f_s(\x^t))).
\end{equation}

As theoretically investigated in \cite{kumar2020understanding}, the target risk with $\f_t$ gets higher as the domain discrepancy between the source domain $\mathcal{P}$ and target domain $\mathcal{Q}$ gets larger. In other words, labeled source data and the corresponding source model are not reliable for the use of distancing target domains. 
This motivates us to introduce reliable auxiliary models of intermediate domains, which could bridge the source and target domains.

\subsection{Self-training of Auxiliary Models} \label{Sec:auxselftrain}

We introduce auxiliary models via self-training. As justified in \cite{kumar2020understanding}, given two datasets with small distribution divergence and a reliable model on one dataset, we could obtain a reliable model for the other one via self-training. Therefore, reliable auxiliary models could be achieved via self-training if intermediate domains, where the domain divergence between consecutive domains is small, are available. In the literature \cite{xie2015transfer,bobu2018adapting,kumar2020understanding}, the expected evolving intermediate domains are assumed to be readily available, which is usually not met in practice. To this end, we propose the following strategy to generate such expected intermediate domains.


\noindent\textbf{Intermediate domains introduction} 
Inspired by \cite{chopra2013dlid}, we start from the source data $\mathcal{S}$, generate successive intermediate datasets by gradually increasing the proportion of samples drawn from $\mathcal{T}$ and decreasing the proportion of samples drawn from $\mathcal{S}$, and end with the target dataset $\mathcal{T}$. Specifically, we introduce $M+1$ datasets $\{ \mathcal{M}_m\}_{m=0}^{M} $, where $\mathcal{M}_0 = \mathcal{S}$ and $\mathcal{M}_M = \mathcal{T} $; the $\mathcal{M}_m$, where $1 \leq m \leq (M-1)$, is composed of the  $\frac{m}{M}$ proportion of the target data $\mathcal{T}$ and $\frac{M-m}{M}$ proportion of the source data $\mathcal{S}$.
Chopra \emph{et al.} \cite{chopra2013dlid} randomly select these samples from $\mathcal{T}$ and $\mathcal{S}$.
We argue that selecting which part of $\mathcal{T}$ or $\mathcal{S}$ may make a difference, since there may be an intra-domain divergence in one dataset  \cite{mancini2018boosting,chen2019blending}. In other words, a specific part of target data may have smaller domain shift to source data, and vice verse. 
Thus we propose the following sample selection strategies and justify their effectiveness in Section \ref{Sec:experiment}.

\noindent\textbf{Target samples selection} 
To minimize the domain divergence between the expected $\mathcal{M}_{m+1}$ and the previous $\mathcal{M}_{m}$, target samples with smaller domain shift to the $\mathcal{M}_{m}$ should be selected for $\mathcal{M}_{m+1}$. 
Inspired by the analysis in Section \ref{Sec:motivation} that samples with smaller domain shift to the labeled data tend to present higher prediction probability based on the model trained with the labeled data, we could achieve this goal by selecting the $\frac{m+1}{M}$ proportion of the target data $\mathcal{T}$ with higher prediction probability based on the $\f_m$, which  is trained with the $m$-th intermediate dataset $\mathcal{M}_{m}$ and will be presented shortly. 
The maximum prediction probability of a sample $\x_i^t$ is:
\begin{equation} \label{Equ:_target_score}
c_t(\x_i^t, \f_m) = \max(\f_m(\x_i^t)).
\end{equation}
Given the prediction probability of all unlabeled target data $\mathcal{C}_t = \{ c_t(\x_i^t, \f_m) \}_{i=1}^{n_t}$, we introduce the binary indicator $v_i^t \in \{ 0, 1\}$ to indicate whether the sample $\x_i^t$ should be selected for the $\mathcal{M}_{m+1}$:
\begin{equation} \label{Equ:target_selection}
v_i^t  =
\begin{cases} 
1, & c_t(\x_i^t, \f_m) \in TOP_{(m+1)n_t/M}(\mathcal{C}_t) \\
0, & Otherwise
\end{cases},
\end{equation}
where $TOP_{N} (\mathcal{C}_t)$ is the subset with $N$ highest scores in $\mathcal{C}_t$.

\noindent\textbf{Source samples selection} 
Given the target domain $\mathcal{T}$ as the terminal of intermediate domains, source data with larger domain shift to the target one should be dropped first. In other words, the source data with smaller domain shift to the target one should be selected for the intermediate domains. Similar to the selection strategy of target samples, we could achieve this goal by introducing a target classifier and selecting source data with higher prediction probability on this classifier.


To introduce the target classifier, we instantiate the auxiliary model $\f_m$ as a standard DNN with the decomposed feature extractor $\phi_m: \mathcal{X} \to \mathcal{Z}$ and the classifier $\varphi_m: \mathcal{Z} \to [0,1]^K$, where $\mathcal{Z}$ is the feature space. 
Based on the auxiliary model $\f_m$, we could introduce a target prototype classifier \cite{snell2017prototypical} with the prototype $\p_k$ of the $k$-th target category as:
\begin{equation} \label{Equ:target_prototype}
\p_k = \frac{\sum_{i=1}^{n_t} \mathbf{1}_{o(\f_m(\x_i^t)) = k}  \phi_m(\x_i^t) }{\sum_{i=1}^{n_t} \mathbf{1}_{o(\f_m(\x_i^t)) = k}  }, 
\end{equation}
where $\mathbf{1}_{y = k} = \begin{cases} 
1, & y = k \\
0, & Otherwise
\end{cases}.$

\begin{figure*}
	\centering
	\subfigure[$\mathcal{M}_0$] {
		\label{Fig:m1}     
		\includegraphics[width=0.185\textwidth]{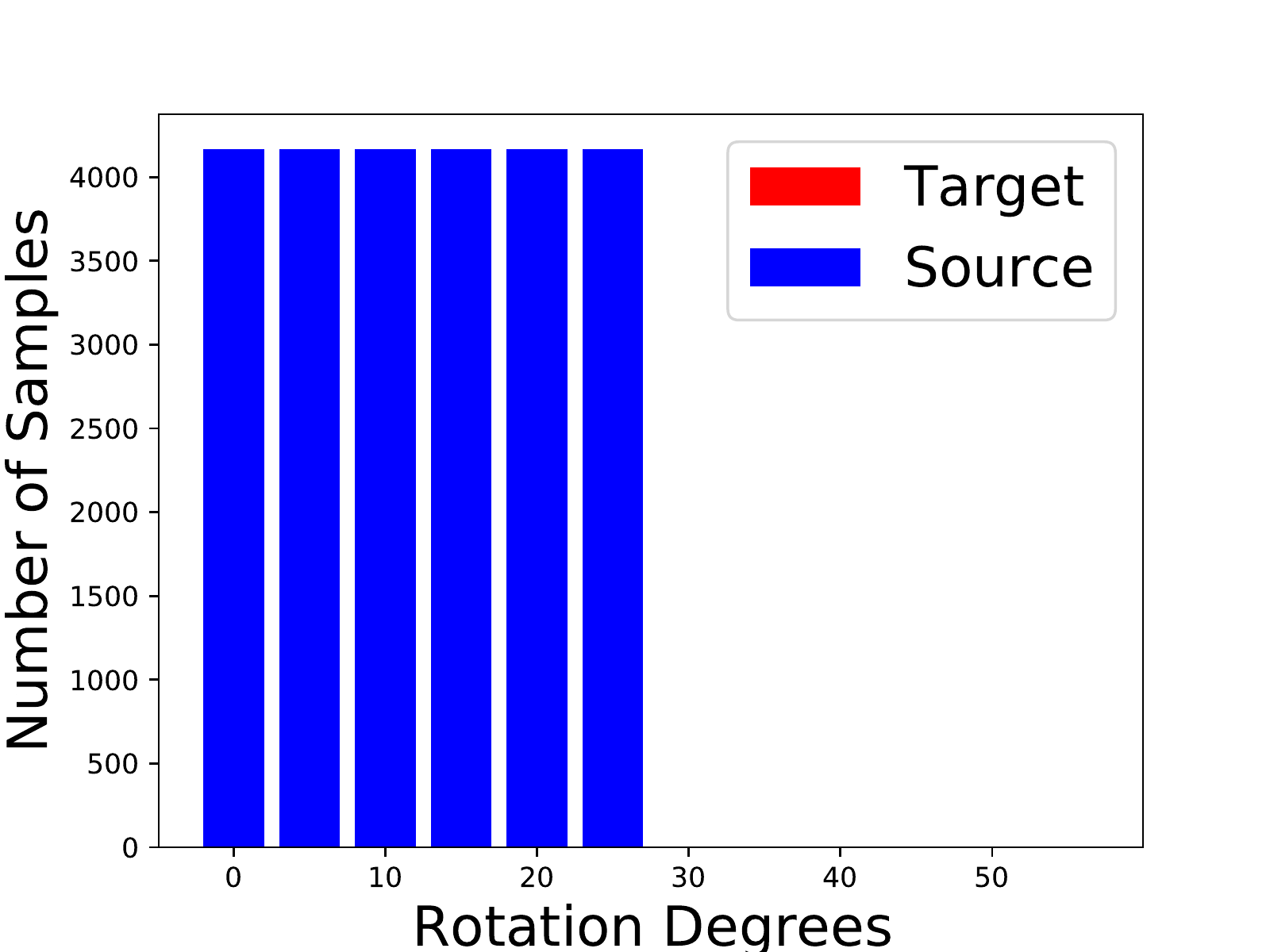}
	} \hfill    
	\subfigure[$\mathcal{M}_{6}$] {
		\label{Fig:scores_vs_shift_mnist}     
		\includegraphics[width=0.185\textwidth]{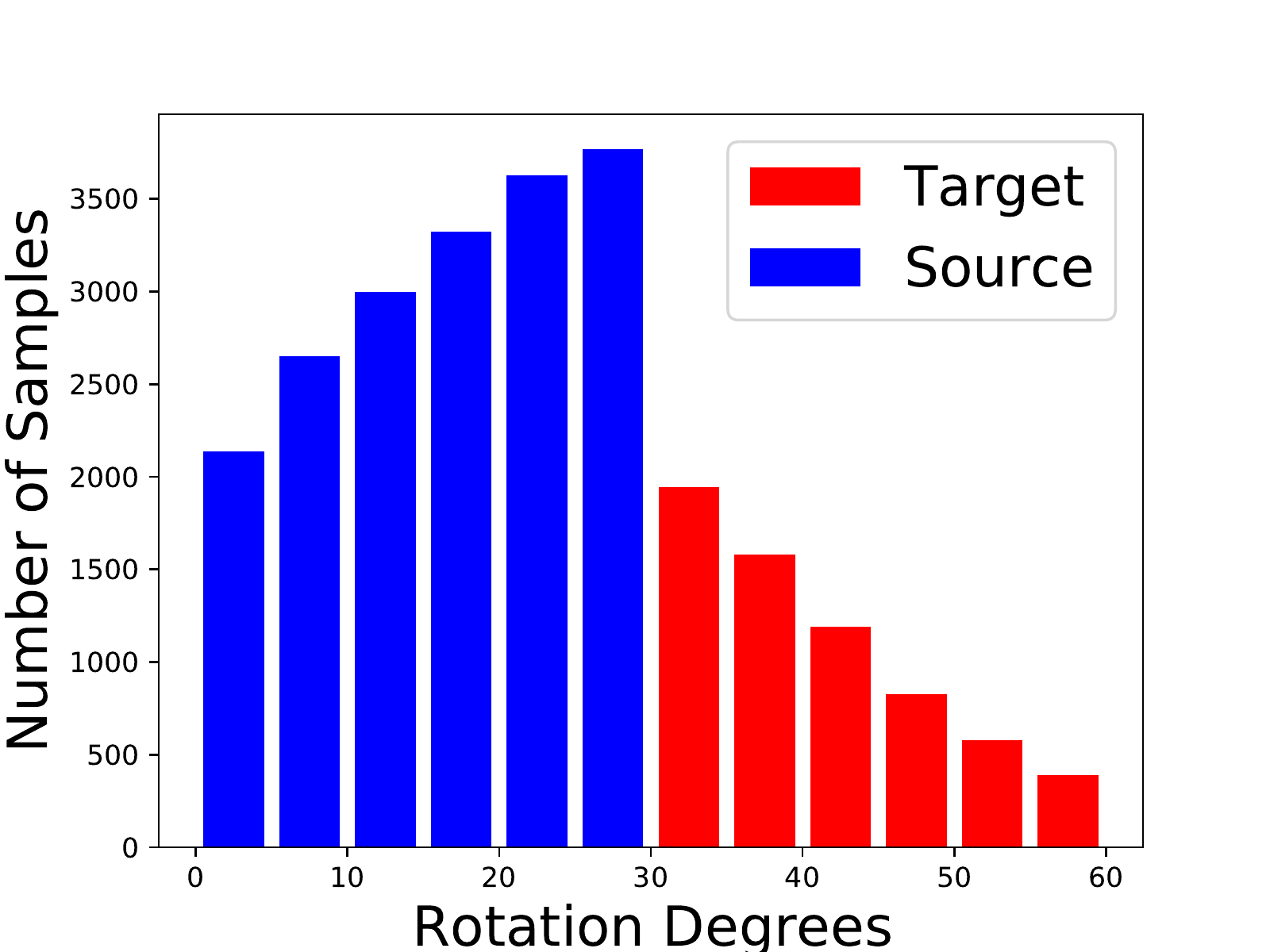}
	}    
	\subfigure[$\mathcal{M}_{12}$] {   
		\label{Fig:linear_process} \includegraphics[width=0.185\textwidth]{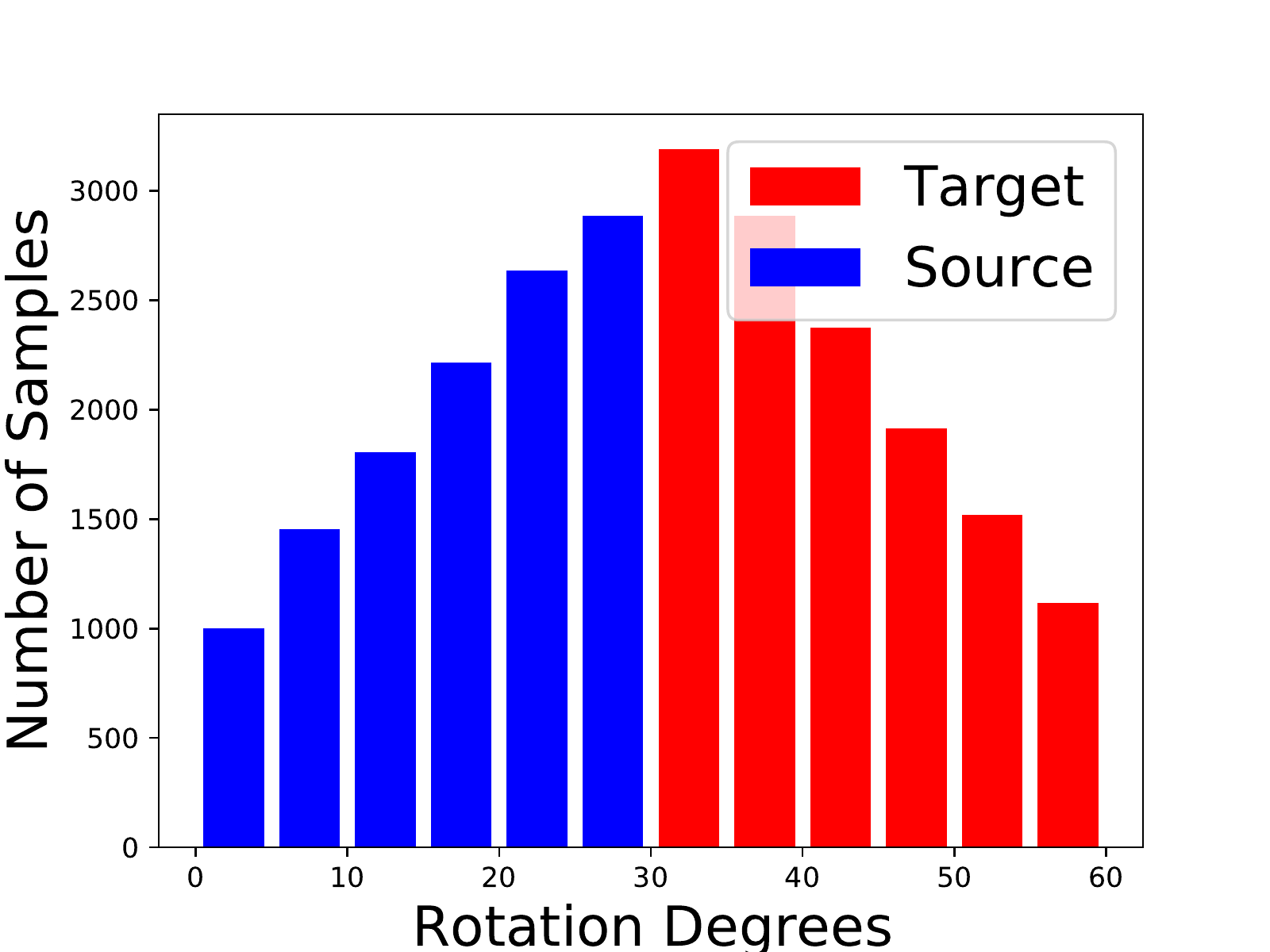}
	} \hfill    
	\subfigure[$\mathcal{M}_{18}$] {
		\includegraphics[width=0.185\textwidth]{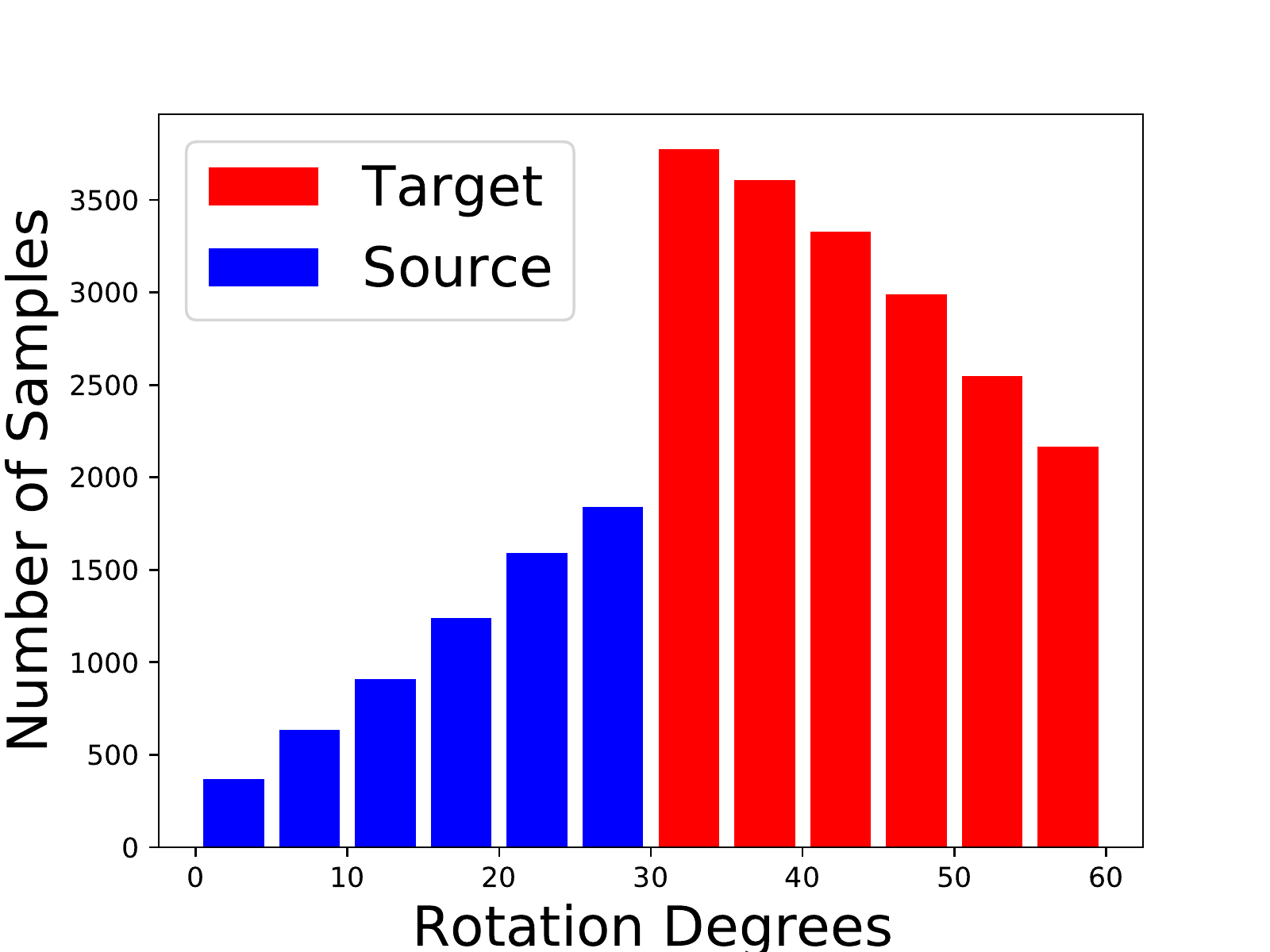} }
	\hfill   
	\subfigure[$\mathcal{M}_{24}$] {
		\includegraphics[width=0.185\textwidth]{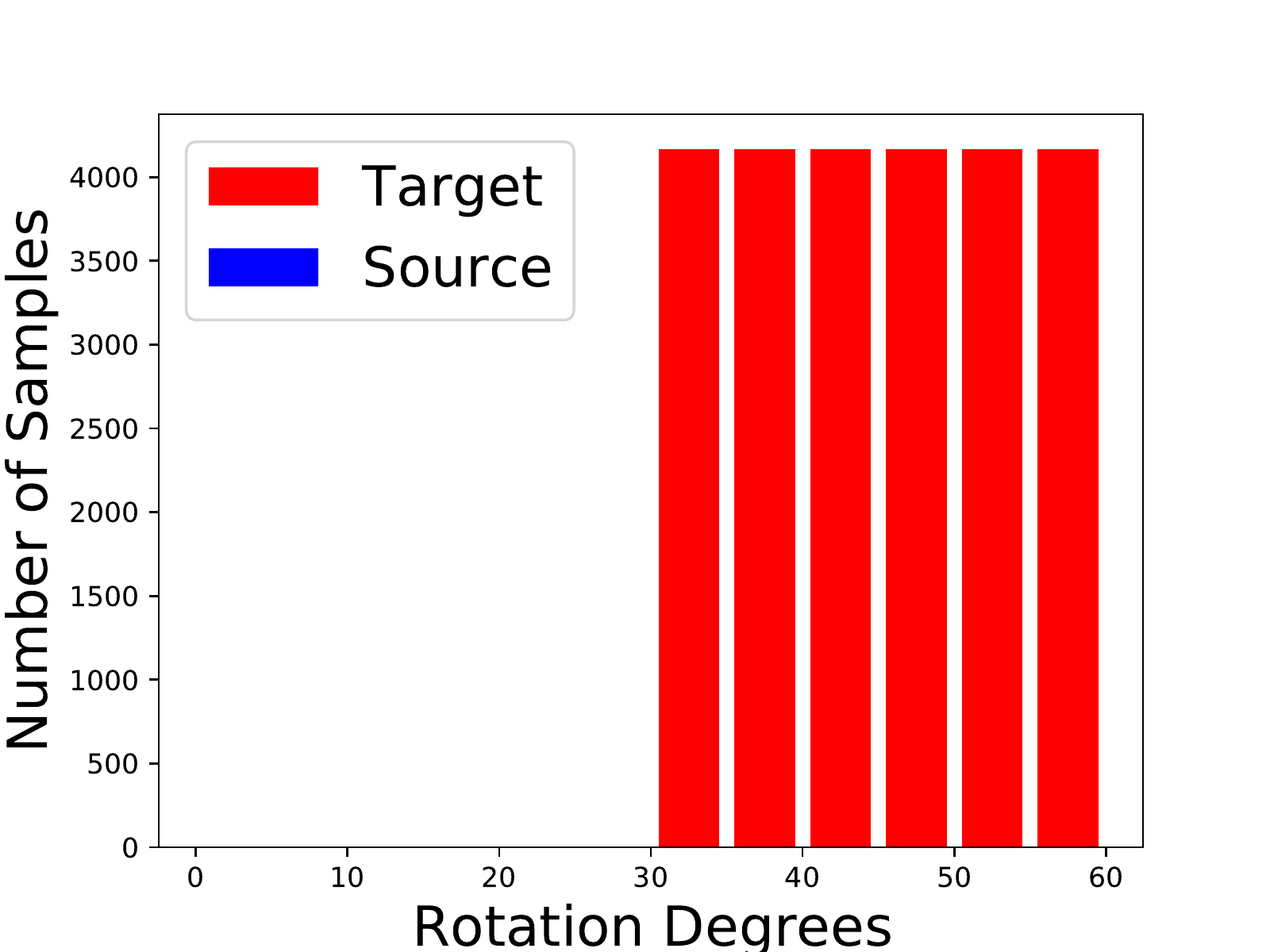} }
	\caption{An illustration of the intermediate domains learned by our AuxSelfTrain on the Rotating MNIST dataset, where we set $M=24$. 
		We randomly rotate source samples by an angle from $0$ to $30$ degrees, while the target samples are randomly rotated from $30$ to $60$ degrees.}   \label{Fig:intermediate_domains}
\end{figure*}

Given target prototypes $\{ \p_k \}_{k=1}^K$, we calculate the prediction probability of a labeled source sample $(\x^s_i, y^s_i)$ as:
\begin{equation} \label{Equ:source_score}
c_s(\x^s_i, \f_m) = 
\frac{\exp(1 + || \phi_m(\x^s_i) - \p_{y^s_i} ||^2)^{-1}}{\sum_{k=1}^{K} \exp(1 + || \phi_m(\x^s_i) - \p_{k} ||^2)^{-1}}.
\end{equation}
Given the prediction probability of all labeled source data $\mathcal{C}_s = \{ c_s(\x^s_i, \f_m) \}_{i=1}^{n_s}$, we also introduce a binary indicator $v_i^s \in \{0,1 \}$ to indicate whether the $\x_i^s$ should be selected to construct the $\mathcal{M}_{m+1}$ as:
\begin{equation} \label{Equ:source_selection}
v_i^s  =
\begin{cases} 
1, & c_s(\x_i^s, \f_m) \in TOP_{(M-m-1)n_s/M}(\mathcal{C}_s) \\
0, & Otherwise
\end{cases}.
\end{equation}
The effectiveness of the proposed sample selection strategies (i.e., (\ref{Equ:target_selection}) and (\ref{Equ:source_selection})) is intuitively presented in Figures \ref{Fig:intermediate_domains} and \ref{Fig:domain_shift_consecutive}, and empirically ablated in Table \ref{Tab:ablation}.

\noindent\textbf{Auxiliary models via self-training}
Given the auxiliary model $\f_m$ and the $(m+1)$-th intermediate dataset $\mathcal{M}_{m+1}$ defined with the target indicator $\{ v_i^t \}_{i=1}^{n_t}$ and the source indicator $\{ v_i^s \}_{i=1}^{n_s}$, the auxiliary model $\f_{m+1}$ could be achieved with the following self-training objective: 
\begin{align} \label{Equ:overall_objective}
\notag \f_{m+1} = \argmin_{\f \in \mathcal{F}} &  \frac{1}{n_s}\sum_{i=1}^{n_s} v_i^s \mathcal{L}(\f(\x^s_i), y^s_i)  \\
+& \frac{1}{n_t}\sum_{i=1}^{n_t} v_i^t \mathcal{L}(\f(\mathcal{A}(\x^t_i)), o(\f_m(\x^t_i))),
\end{align}
where we utilize the advanced data augmentation $\mathcal{A}(\cdot)$ \cite{cubuk2020randaugment} for unlabeled data following \cite{sohn2020fixmatch}.
We initialize $\f_0 = \f_s$ (\ref{Equ:source_pre_training}) as the model trained on the labeled source data. 
Note that training each auxiliary model from scratch is time consuming. However, given evolving domains, minor computation is required to adapt the learned $\f_m$ to $\f_{m+1}$. In this way, our method only doubles the training iterations of the Source Only baseline (including the training of $\f_s$ (\ref{Equ:source_pre_training})), leading to an efficient implementation.




\subsection{An Enhanced Indicator for Sample Selection}
Although the maximum prediction probability (\ref{Equ:_target_score}) is a valid indicator of domain shift, as showed in Figure \ref{Fig:score_acc_dis_vs_shift}, it is possible that unlabeled samples with high prediction probability may be distant from labeled ones, since DNNs may produce high probability for inputs far away from training data \cite{nguyen2015deep}.  To this end, we propose an enhanced indicator for sample selection via implicit ensemble. 

As presented in \cite{lakshminarayanan2016simple,ovadia2019can} with a large scale empirical study, ensemble is a simple but efficient way to improve the quality of predictive uncertainty to domain shift. However, vanilla ensemble by training multiple models is computational expensive and memory consuming \cite{lakshminarayanan2016simple,sagi2018ensembles,ovadia2019can}. Therefore, we propose an  implicit ensemble scheme by introducing two accessory models $\widehat{\f}_m$ and $\widetilde{\f}_m$, where all three models (i.e., $\f_m$, $\widehat{\f}_m$, and $\widetilde{\f}_m$) share the same feature extractor $\phi_m$; $\widehat{\f}_m$ and $\widetilde{\f}_m$ are parameter-free and computational efficient. The enhanced indicator is introduced as:
	\begin{equation} \label{Equ:f_refinement}
	\f_m(\x^t) = (\f_m(\x^t) + \widehat{\f}_m(\x^t) + \widetilde{\f}_m(\x^t) ) /3. 
	\end{equation}  
	Note that the enhanced indicator $\f_m$  (\ref{Equ:f_refinement}) is only adopted in training process (i.e., $\f_m$ in (\ref{Equ:target_selection}) and (\ref{Equ:overall_objective})), and we still use the vanilla $\f_m$ (i.e., without implicit ensemble) for the test. The details of $\widehat{\f}_m$ and $\widetilde{\f}_m$ are given as follows.

The $\widehat{\f}_m$ is a classifier constructed from the clustering method, which is widely adopted in DA \cite{can,srdc}. We first calculate source category centers $\{ \mathbf{O}_k^s \}_{k=1}^K$ as: $\mathbf{O}_k^s = \frac{1}{\sum_{j=1}^{n_s}  \mathbf{1}_{y_{j}^s = k}} \sum_{i=1}^{n_s} \mathbf{1}_{y_{i}^s = k} \phi_m(\x_i^s)$. Then we apply the k-means clustering on extracted target features $\{ \phi_m(\x_i^t) \}_{i=1}^{n_t}$, and use source category centers $\{ \mathbf{O}_k^s \}_{k=1}^K$ as initial cluster centers.  After clustering, we could get the corresponding target cluster centers $\{ \mathbf{O}_k^t \}_{k=1}^K$. The $\widehat{\f}_m$ is thus introduced with its $k$-th entry $\widehat{f}_{m,k} (\x^t)$ for any target sample $\x^t$ as:
\begin{equation} \label{Equ:clustering}
\widehat{f}_{m,k}(\x^t) = \frac{\exp(1 + || \phi(\x^t) - \mathbf{O}_k^t ||^2)^{-1}}{\sum_{k'=1}^{K} \exp(1 + || \phi(\x^t) - \mathbf{O}_{k'}^t ||^2)^{-1}}.
\end{equation}

The $\widetilde{\f}_m$ is a classifier based on the label propagation \cite{zhou2004learning}, which is also widely adopted in DA \cite{ding2018graph,zhang2020label}. 
We could achieve $\widetilde{\f}_m$ by minimizing the following objective:
\begin{equation} \label{Equ:label_propgation}
\sum_{i=1}^{n_{st}} \| \widetilde{\f}_m (\x_i^{st}) - \y_i^{st} \|^2 + \lambda \sum_{i,j}^{n_{st}} a_{ij} \| \frac{\widetilde{\f}_m(\x_i^{st})}{\sqrt{d_{ii}}}   - \frac{\widetilde{\f}_m(\x_j^{st})}{\sqrt{d_{jj}}} \|^2,
\end{equation} 
where $n_{st} = n_s+n_t$, $\x_i^{st} = \begin{cases}
\x_i^s & i \leq n_s \\
\x_{i-n_s}^t & others \\
\end{cases}, \y_i^{st}$ is a one-hot vector of $K$ dimension with the $y_i^s$-th entry of $1$ if $i \leq n_s$ otherwise $\y_i^{st} = \{0\}^K$, 
the cosine similarity $a_{ij} = \frac{<\phi_m(\x_i^{st}), \phi_m(\x_j^{st})>}{ \| \phi_m(\x_i^{st})\|   \|\phi_m(\x_j^{st})\|}$ between features $\phi_m(\x_i^{st})$ and $\phi_m(\x_j^{st})$ is the element of the affinity matrix $\mathbf{A} \in \mathbb{R}_{+}^{{n_{st}} \times {n_{st}}}$ and $d_{ii}$ is the sum of $i$-th row of $\mathbf{A}$. The solution of (\ref{Equ:label_propgation}) is given in the appendices. 
We normalize the output of $\widetilde{\f}_m$ as $\widetilde{\f}_m (\x^t) = \widetilde{\f}_m (\x^t) / \sum \widetilde{\f}_m (\x^t) $ for the use in (\ref{Equ:f_refinement}).

Given pre-trained feature extractor $\phi_m$ of a auxiliary model $\f_m$ (\ref{Equ:overall_objective}), accessory models $\widehat{\f}_m$ and $\widetilde{\f}_m$ could be achieved immediately and the diversity of ensemble members is naturally satisfied via different algorithms. 
Although better classification accuracy may be also introduced via implicit ensemble, we find that the performance improvement is mainly from the improved sample selection via improved quality of predictive uncertainty to domain shift, as illustrated in Table \ref{Tab:ablation}.
The AuxSelfTrain algorithm is illustrated in Algorithm \ref{alg:framework} and intuitively presented in Figure \ref{Fig:framework}.

\begin{algorithm}
	\caption{Algorithm Framework of AuxSelfTrain.} \label{alg:framework}
	\begin{algorithmic}[1]
		\Require Source data $\mathcal{S} = \{ \x_i^s, y_i^s \}_{i=1}^{n_s}$, target data $\mathcal{T}= \{ \x_i^t \}_{i=1}^{n_t}$, number of intermediate domains $M$
		\State Initialize $\f_0$ as source model $\f_s$ (\ref{Equ:source_pre_training})
		\For{$m=1$; $m\leq M$; $m++$}
		\State Acquire the enhanced $\f_{m-1}$ using (\ref{Equ:f_refinement})
		\State Acquire intermediate dataset $\mathcal{M}_{m}$ via (\ref{Equ:target_selection}) and (\ref{Equ:source_selection})
		\State Acquire the auxiliary model $\f_m$ using (\ref{Equ:overall_objective})
		\EndFor
		\State Adopt $\f_M$ for testing on the target domain
	\end{algorithmic}
\end{algorithm}


\subsection{Extension to SSDA}
Our AuxSelfTrain could be easily extended to SSDA tasks, where a few target data $\{ \x_i^{t} \}_{i=1}^{n_l}, n_l \ll n_t$, are manually assigned with category labels   $\{ y_i^{t} \}_{i=1}^{n_l}$. 
As illustrated in Section \ref{Sec:auxselftrain}, we introduce intermediate datasets as combinations of source and target samples. Therefore, the combination of labeled source data and labeled target data in SSDA could be treated as a pre-given intermediate domain. We could start our AuxSelfTrain from this intermediate domain and replace $o(\f_m(\x_i^t))$ with $y_i^{t}$, $1\leq i \leq n_l$,  to utilize manual labels. Empirical results in Section \ref{Sec:experiment} justify its efficacy.

\section{Experiments} \label{Sec:experiment}
We first conduct control studies on the Rotating MNIST dataset. Then we evaluate our AuxSelfTrain on standard benchmark datasets of DA and SSDA.

\subsection{Control Studies} \label{Sec:control_study}

Rotating MNIST is a semi-synthetic dataset from the MNIST \cite{mnist}, where the rotation degree of each sample is strictly controlled. Specifically, we randomly split the $50,000$ training samples of the MNIST into the source and target domains of equal size. The source samples are randomly rotated by an angle from $0$ to $30$ degrees to construct the labeled source domain. We construct the unlabeled target domain by randomly rotating target samples with degrees of $30\sim 60$. 
A $3$-layer convolutional network with dropout and batch normalization is adopted. 

%
%
%

\begin{figure}
	\centering
	\subfigure[$\mathcal{A}_{dis}$] {
		\label{Fig:domain_shift_mnist}
		\includegraphics[width=0.22\textwidth]{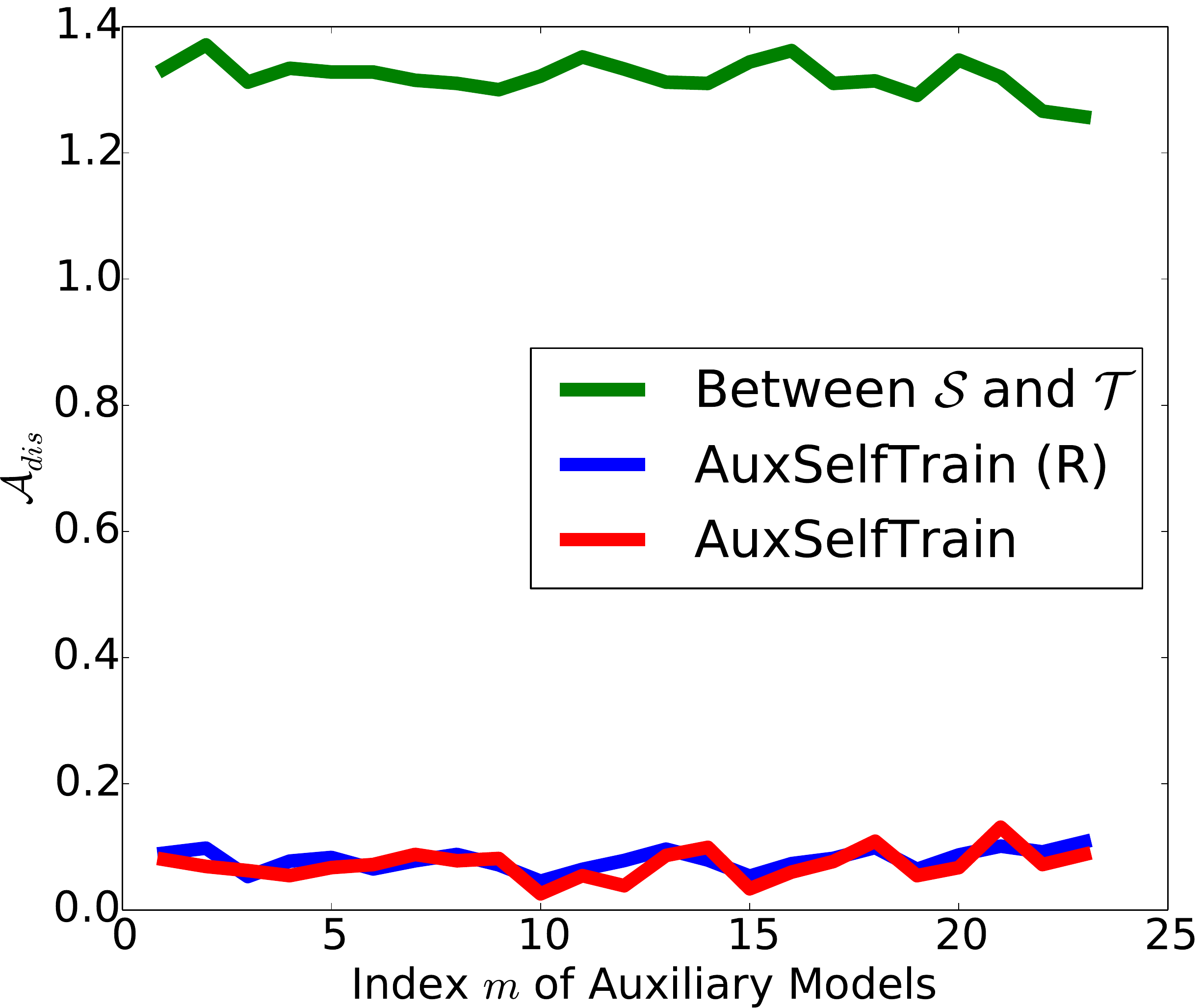}
	} \hfill    
	\subfigure[W$_{\infty}$-based discrepancy] {
		\label{Fig:wass_disc}
		\includegraphics[width=0.22\textwidth]{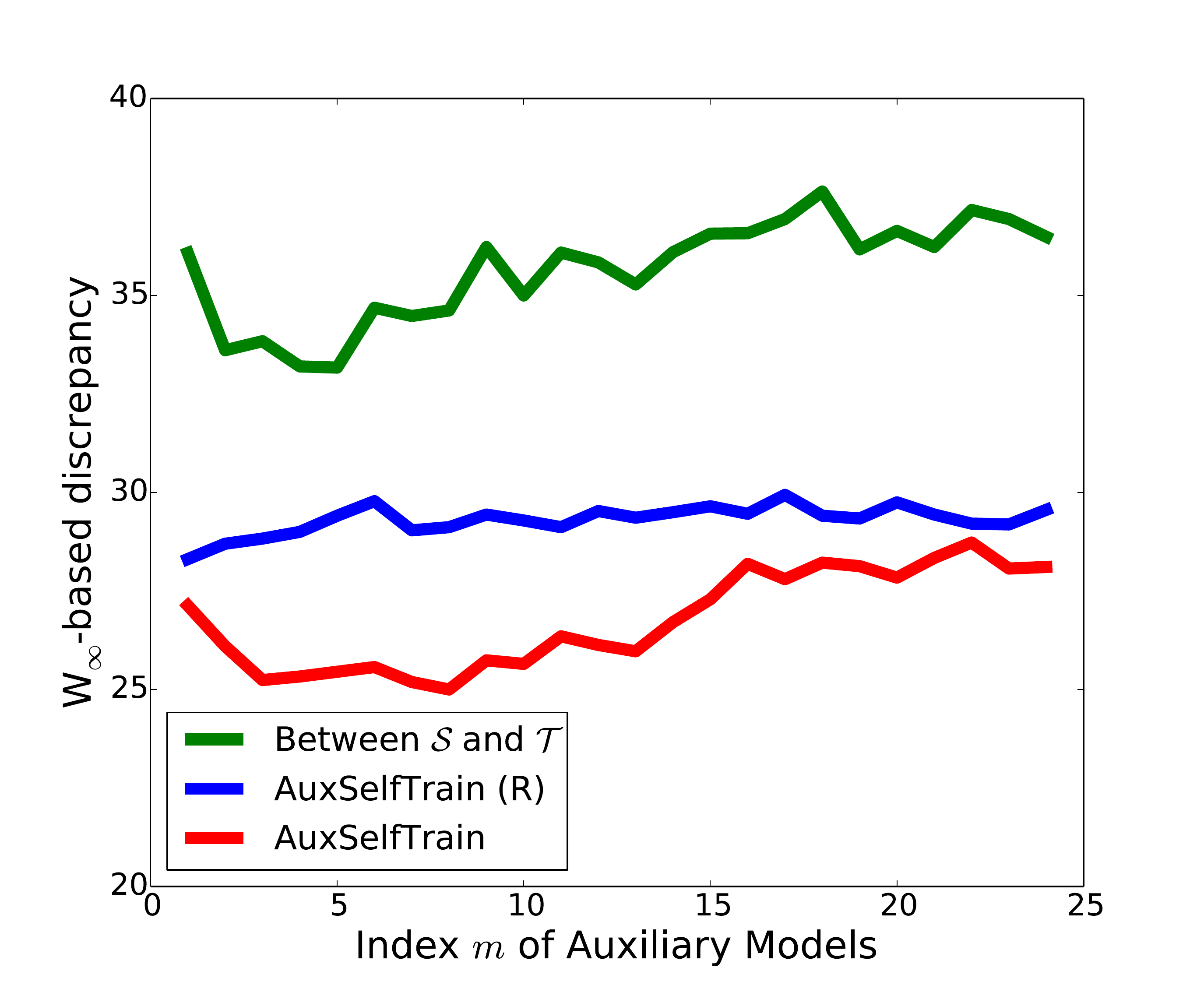} }
	\caption{Illustrations of (a) $\mathcal{A}_{dis}$ \cite{ben2007analysis}  and (b) W$_{\infty}$-based discrepancy \cite{kumar2020understanding} on Rotating MNIST dataset based on feature representations of $m$-th auxiliary model $\f_m$. The ``Between $\mathcal{S}$ and $\mathcal{T}$'' indicate the discrepancy across source and target domains. The AuxSelfTrain and AuxSelfTrain (R) represent the discrepancy between $\mathcal{M}_m$ and $\mathcal{M}_{m-1}$ in corresponding methods. Following \cite{chopra2013dlid}, we randomly sample instances to construct intermediate domains in AuxSelfTrain (R). }   \label{Fig:domain_shift_consecutive}
\end{figure}

\begin{table*}
	\centering
	\begin{tabular}{cccc|cccccccccccccc}
		\hline
		 \rotatebox{90}{Selection-T} & \rotatebox{90}{Selection-S} & \rotatebox{90}{Selection-Enh.} & \rotatebox{90}{Labeling-Enh.} &  \rotatebox{90}{plane} & \rotatebox{90}{bicycle} & \rotatebox{90}{bus} & \rotatebox{90}{car} & \rotatebox{90}{horse} & \rotatebox{90}{knife} & \rotatebox{90}{motorcycle} & \rotatebox{90}{person} & \rotatebox{90}{plant} & \rotatebox{90}{skateboard} & \rotatebox{90}{train} & \rotatebox{90}{truck} & \rotatebox{90}{mean} \\
		 \hline
		 $\checkmark$ & $\checkmark$ & $\checkmark$ & $\checkmark$   & \textbf{95.0} & \textbf{86.9} & 84.8 & 80.4 & 95.5 & \textbf{90.5} & 93.1 & \textbf{84.6} & \textbf{95.1} & 85.6 & 89.8 & \textbf{40.6} & \textbf{85.2} \\
		 \hline
		 $\times$ & $\checkmark$  & $\checkmark$ & $\checkmark$ & 88.8 & 65.0 & 73.8 & 73.0 & 93.1 & 61.9 & 80.8 & 71.0 & 91.9 & 67.0 & 51.9 & 20.4 & 69.9    \\
		 R & $\checkmark$  & $\checkmark$ & $\checkmark$ & 93.8 & 74.8 & 75.8 & 72.6 & 94.7 & 82.0 & 75.1 & 81.5 & 93.1 & 84.0 & 57.9 & 29.8 & 76.3    \\
		 $\checkmark$ & $\times$  & $\checkmark$ & $\checkmark$   & \textbf{95.0} & 84.1 & 85.9 & \textbf{81.1} & 95.2 & 80.1 & 92.8 & 83.8 & 94.9 & 80.4 & 88.6 & 31.8 & 82.8  \\
		 $\checkmark$ & R  & $\checkmark$ & $\checkmark$   & \textbf{95.0} & 85.4 & 85.2 & \textbf{81.1} & 95.4 & 88.3 & 93.1 & \textbf{84.6} & 94.7 & \textbf{85.9} & 89.3 & 38.3 & 84.7   \\
		 $\checkmark$ & $\checkmark$  & $\times$ & $\times$  & 93.9 & 85.1 & \textbf{87.3} & 83.2 & \textbf{95.8} & 74.0 & \textbf{93.4} & 84.5 & 93.3 & 77.9 & 90.1 & 39.4 & 83.2  \\
		 $\checkmark$ & $\checkmark$  & $\checkmark$ & $\times$  &   94.3 & 85.8 & 85.5 & 80.9 & 95.7 & 86.5 & 93.1 & 83.6 & 93.3 & \textbf{85.9} & \textbf{90.7} & 39.7 & 84.6 \\ 
		\hline

		\hline
	\end{tabular}
	\caption{Ablation experiments on VisDA-2017 dataset (ResNet50).  The $\checkmark$, R, and $\times$ symbols in the `Selection-T' column indicate adopting our target selection scheme (\ref{Equ:target_selection}), randomly selecting the same number of target data as our target selection scheme, and selecting all target data (i.e., $v_i^t=1, 1\leq i \leq n_t$), respectively. The $\checkmark$, R, and $\times$ symbols in the `Selection-S' column are similarly defined. The $\checkmark$ and $\times$ symbols in the `Selection-Enh.' column indicate using the enhanced indicator (\ref{Equ:f_refinement}) and the vanilla $\f_m$ in the sample selection scheme (\ref{Equ:target_selection}), respectively; the $\checkmark$ and $\times$ symbols in the `Labeling-Enh.' column indicate using the enhanced indicator (\ref{Equ:f_refinement}) and the vanilla $\f_m$ to provide pseudo labels of target data (i.e., $o(\f_m(\x_i^t))$ in (\ref{Equ:overall_objective})), respectively.  }
	\label{Tab:ablation}
\end{table*}

To illustrate the effectiveness of our method intuitively, we present the sample distribution in different intermediate domains learned by our AuxSelfTrain. As intermediate domains evolve from the source domain $\mathcal{M}_0$ to target one $\mathcal{M}_M$, source samples with smaller rotation degrees, which are far away from target domain, are firstly dropped and target samples with smaller rotation degrees are firstly selected, as illustrated in Figure \ref{Fig:intermediate_domains}. This makes the domain divergence between consecutive domains small although the divergence between source and target domains is large, as showed in Figure \ref{Fig:domain_shift_consecutive};
the evolving intermediate domains support the successful self-training in AuxSelfTrain \cite{kumar2020understanding}.

\subsection{Analyses and Results} \label{Sec:Results}

\noindent  \textbf{VisDA-2017} \cite{peng2017visda} is a dataset with large domain shift from synthetic data (\textbf{Syn.}) to real images (\textbf{Real}), which contains about $280K$ images across $12$ categories. 
\textbf{OfficeHome} \cite{home} is a standard benchmark dataset of DA, which contains $65$ classes shared by $4$ domains; we adopt it in both DA and SSDA experiments. 
\textbf{DomainNet} \cite{peng2019moment} is a large-scale benchmark dataset, containing $345$ classes shared by six domains and about $0.6$ million images. 
Following \cite{ssda_mme}, we conduct SSDA on a subset of DomainNet, where $126$ classes are selected and $7$ tasks are defined. 
We use all labeled source data and all unlabeled target data for training in DA. In SSDA, we adopt the split strategy in  \cite{ssda_mme}, where one or three target data are manually labeled per class.

\noindent \textbf{Implementation details}  We conduct experiments based on the ResNet model \cite{resnet} and VGG16 model \cite{vgg} using Pytorch. 
Given a model pre-trained on ImageNet dataset, we remove the last fully connect (FC) layer and construct the feature extractor $\phi(\cdot)$ with remaining parts; a new FC layer with $K$ neurons is adopted as the classifier $\varphi(\cdot)$. We sample two batches with different sizes in each iteration; the first batch contains $B$ labeled data, which are sampled from the labeled source data in DA and from the labeled source and labeled target data in SSDA. 
Following \cite{xie2020unsupervised,sohn2020fixmatch}, we sample a large batch of unlabeled data with batch size of $u B$, where $u \ge 1$; we set $B=64$ and $u=7$ in all experiments.
All parameters are updated by SGD with momentum of 0.9. Following \cite{dann}, we set the learning rate as: $\eta_p = \frac{\eta_0}{(1+\alpha p)^{\beta}}$, where $\eta_0=0.01$, $\alpha=10$, $\beta=0.75$, and $p$ is linearly changing from 0 to 1 as the training proceeds. 

\begin{table}
	\centering
	\begin{tabular}{l|c}
		\hline
		Methods & Syn. $\to$ Real (\%) \\
		\hline
		Source Only \cite{resnet}  & 45.6 \\
		DANN \cite{dann} & 55.0 \\
		MCD \cite{mcd} & 69.8\\
		MDD \cite{zhang2019bridging} & 74.6 \\
		GSDA \cite{hu2020unsupervised} & 81.5 \\
		RWOT \cite{xu2020reliable} & 84.0 \\
		\hline
		AuxSelfTrain & \textbf{85.2} \\
		\hline
	\end{tabular}
	\caption{Results (\%) on the VisDA-2017 dataset for the DA task (ResNet50). } \label{Table:visda}
\end{table}

\begin{table*}[h!]\small
	\begin{center}
		\begin{tabular}{L{23.0mm}C{7.7mm}C{7.7mm}C{7.7mm}C{7.7mm}C{7.7mm}C{7.7mm}C{7.7mm}C{7.7mm}C{7.7mm}C{7.7mm}C{7.7mm}C{7.7mm}C{7.7mm}C{7.9mm}}
			\hline
			Methods                  &Ar$\to$Cl &Ar$\to$Pr &Ar$\to$Rw &Cl$\to$Ar &Cl$\to$Pr &Cl$\to$Rw &Pr$\to$Ar &Pr$\to$Cl &Pr$\to$Rw &Rw$\to$Ar &Rw$\to$Cl &Rw$\to$Pr    & Avg  \\
			\hline
			Source Only \cite{resnet}   &34.9      & 50.0     & 58.0      & 37.4      & 41.9      & 46.2     & 38.5     & 31.2     & 60.4     & 53.9     & 41.2     & 59.9 &46.1 \\		
			DANN \cite{reverse_grad,dann}& 45.6     & 59.3     & 70.1      & 47.0      & 58.5      & 60.9     & 46.1     & 43.7     & 68.5     & 63.2     &51.8      & 76.8 &57.6 \\		
			
			SymNets \cite{symnets}& 47.7 & 72.9          & 78.5            & 64.2          &71.3          &74.2 &64.2          &48.8 &79.5          &74.5  &52.6 &82.7          & 67.6  \\
			MDD     \cite{zhang2019bridging}     & 54.9    & 73.7 &77.8 & 60.0 & 71.4 & 71.8 &61.2 & 53.6 & 78.1 & 72.5 & 60.2 & 82.3 & 68.1 \\		
			HDAN \cite{cui2020hda} &  56.8 & 75.2 & 79.8 & 65.1 & 73.9 & 75.2 & 66.3 & 56.7 & 81.8 & 75.4 & 59.7 & 84.7 & 70.9 \\
			GSDA \cite{hu2020unsupervised} & \textbf{61.3} & 76.1 & 79.4 & 65.4 & 73.3 & 74.3 & 65.0 & 53.2 & 80.0 & 72.2 & 60.6 & 83.1 & 70.3 \\ 
			SRDC \cite{srdc} & 52.3 & 76.3 & 81.0 & \textbf{69.5} & 76.2 & \textbf{78.0} & \textbf{68.7} & 53.8 & 81.7 & \textbf{76.3} & 57.1 & 85.0 & 71.3 \\
			\hline
			AuxSelfTrain & 55.3 & \textbf{81.6} & \textbf{82.4} & 68.8 & \textbf{77.2} & 77.2 & 66.1 & \textbf{57.2} & \textbf{83.0} & 73.0 & \textbf{61.1} & \textbf{86.8} & \textbf{72.5} \\
			\hline
		\end{tabular}  
		\caption{Results (\%) on the OfficeHome dataset \cite{home} for the DA task (ResNet50). }
		\label{Tab:home_close_da}
	\end{center}
\end{table*}

\begin{table*}[htb]\footnotesize
	\begin{center}
		\begin{tabular}{L{9.5mm}|L{16.9mm}|C{7.0mm}C{7.0mm}C{7.0mm}C{7.0mm}C{7.0mm}C{7.0mm}C{7.0mm}C{7.0mm}C{7.0mm}C{7.0mm}C{7.0mm}C{7.0mm}C{7.0mm}}
			\hline
			Network & Methods     &Rw$\to$Cl &Rw$\to$Pr  &Rw$\to$Ar  &Pr$\to$Rw  &Pr$\to$Cl  &Pr$\to$Ar  &Ar$\to$Pr &Ar$\to$Cl &Ar$\to$Rw &Cl$\to$Rw  &Cl$\to$Ar &Cl$\to$Pr    & Avg  \\
			\hline
			\multirow{3}{*}{VGG16} & S+T & 49.6 & 78.6 & 63.6 & 72.7 & 47.2 & 55.9 & 69.4 & 47.5 & 73.4 & 69.7 & 56.2 & 70.4 & 62.9 \\
			& MME \cite{ssda_mme} & 56.9 & 82.9 & 65.7 & 76.7 & 53.6 & 59.2 & 75.7 & 54.9 & 75.3 & 72.9 & 61.1 & 76.3 & 67.6 \\
			\cline{2-15}
			& AuxSelfTrain & \textbf{61.3} & \textbf{85.3} & \textbf{69.5} & \textbf{78.9} & \textbf{58.8} & \textbf{63.6} & \textbf{79.7} & \textbf{59.3} & \textbf{78.1} & \textbf{75.6} & \textbf{64.5} & \textbf{79.8} & \textbf{71.2}\\
			\hline
			
			\hline
			\multirow{4}{*}{ResNet34} & S+T & 55.7 & 80.8 & 67.8 & 73.1 & 53.8 & 63.5 & 73.1 & 54.0 & 74.2 & 68.3 & 57.6 & 72.3 & 66.2 \\
			& MME \cite{ssda_mme} & 64.6 & 85.5 & 71.3 & 80.1 & 64.6 & 65.5 & 79.0 & 63.6 & 79.7 & 76.6 & 67.2 & 79.3 & 73.1 \\
			& Kim \emph{et. al}  \cite{kim2020attract} & 66.4 & 86.2 & 73.4 & 82.0 & \textbf{65.2} & 66.1 & 81.1 & 63.9 & 80.2 & 76.8 & 66.6 & 79.9 & 74.0  \\
			\cline{2-15}
			& AuxSelfTrain & \textbf{66.8} & \textbf{88.0} & \textbf{75.3} & \textbf{83.1} & 64.7 & \textbf{72.1} & \textbf{83.8} & \textbf{65.7} & \textbf{81.4} & \textbf{79.5} & \textbf{70.1} & \textbf{82.7} & \textbf{76.1} \\
			\hline
		\end{tabular}
		\caption{Results (\%)  on the OfficeHome dataset \cite{home} for the SSDA task, where three labeled target data are provided per category. Results with one labeled target data per category are provided in the appendices.}
		\label{Tab:Office-Home-ssda3-vgg}
	\end{center}
\end{table*}

\begin{table*}[h!]\footnotesize
	\centering
	\begin{tabular}{L{19.0mm}C{5.4mm}C{5.4mm}C{5.4mm}C{5.4mm}C{5.4mm}C{5.4mm}C{5.4mm}C{5.4mm}C{5.4mm}C{5.4mm}C{5.4mm}C{5.4mm}C{5.4mm}C{6.4mm}C{6.4mm}C{6.4mm}}
		\hline
		\multirow{2}{*}{Method} & \multicolumn{2}{c}{R$\to$C} & \multicolumn{2}{c}{R$\to$P} & \multicolumn{2}{c}{P$\to$C}  & \multicolumn{2}{c}{C$\to$S}   & \multicolumn{2}{c}{S$\to$P} & \multicolumn{2}{c}{R$\to$S} & \multicolumn{2}{c}{P$\to$R}  & \multicolumn{2}{c}{Avg} \\
		& 1{-s} & 3{-s} & 1{-s} & 3{-s} & 1{-s} & 3{-s} & 1{-s} & 3{-s} & 1{-s} & 3{-s} & 1{-s} & 3{-s} & 1{-s} & 3{-s} & 1{-s} & 3{-s}  \\
		\hline
		S+T  &   55.6 & 60.0  & 60.6  & 62.2 & 56.8 & 59.4 & 50.8 & 55.0  & 56.0 & 59.5  & 46.3  & 50.1  & 71.8  & 73.9  & 56.9 & 60.0 \\ 
		MME\cite{ssda_mme} & 70.0 & 72.2 & 67.7 & 69.7 & 69.0 & 71.7 & 56.3 & 61.8 & 64.8 & 66.8 & 61.0 & 61.9 & 76.1 & 78.5 & 66.4 & 68.9  \\ 
		Meta-MME \cite{li2020online} & -- & 73.5 & -- & 70.3 & -- & 72.8 & -- & 62.8 & -- & 68.0 & -- & 63.8 & -- & 79.2 & -- & 70.1 \\
		BiAT \cite{jiangbidirectional}  & 73.0 & 74.9 & 68.0 & 68.8 & 71.6 & 74.6 & 57.9 & 61.5 & 63.9 & 67.5 & 58.5 & 62.1 & 77.0 & 78.6 & 67.1 & 69.7 \\
		HDAN \cite{cui2020hda} & 71.7 & 73.9 & 67.1 & 69.1 & 72.8 & 73.0 & 63.7 & 66.3 & 65.7 & 67.5 & 69.2 & 69.5 & 76.6 & 79.7 & 69.5 & 71.3 \\ 
		Kim \emph{et. al}  \cite{kim2020attract}                                             & 70.4 & 76.6 & 70.8 & 72.1 & 72.9 & 76.7 & 56.7 & 63.1 & 64.5 & 66.1 & 63.0 & 67.8 & 76.6 & 79.4 & 67.6 & 71.7 \\
		\hline
		AuxSelfTrain & \textbf{78.4} & \textbf{78.8} & \textbf{75.2} &  \textbf{75.4} & \textbf{75.7}& \textbf{77.7} &   \textbf{70.8} &  \textbf{72.0} &  \textbf{73.7} &  \textbf{74.7} & \textbf{72.0} & \textbf{72.1} &  \textbf{83.6} &  \textbf{84.7}  & \textbf{75.6} & \textbf{76.4} \\
		\hline
	\end{tabular}
	\caption{Results (\%)  on the DomainNet dataset for the SSDA task (ResNet34).}
	\label{Tab:domainnet_ssda}
\end{table*}

\subsubsection{Analyses} \label{Sec:analyses}

\noindent \textbf{Ablation studies}
We investigate the target sample selection (\ref{Equ:target_selection}), source sample selection (\ref{Equ:source_selection}) and enhanced indicator (\ref{Equ:f_refinement}) by removing the corresponding component from the overall objective. Specifically, setting $v_i^t=1, 1 \leq i \leq n_t$ in (\ref{Equ:target_selection}) (i.e., $\times$ in `Selection-T' column) leads to a significant performance degradation; replacing our target selection scheme by randomly selecting the same number of target data (i.e., R in `Selection-T' column) achieves worse results than our AuxSelfTrain, although it improves over the setting without sample selection (i.e., $\times$ in `Selection-T' column). The similar results are also observed for the source sample selection (i.e., results with $\checkmark$, R, and $\times$ in the `Selection-S' column). 
These results justify the necessity of  constructing evolving intermediate domains to facilitate self-training and the efficacy of our proposed sample selection strategies.
The enhanced indicator $\f_m$ (\ref{Equ:f_refinement}) plays two roles in our AuxSelfTrain, i.e., providing indicators for target sample selection (\ref{Equ:target_selection}) and generating pseudo labels (i.e., $o(\f_m(\x_i^t))$ in (\ref{Equ:overall_objective})) for target data. To investigate the two roles individually, we keep the enhanced indicator for target sample selection and generate pseudo labels of target data with the non-enhanced $\f_m$ (i.e., $\checkmark$ in `Selection-Enh.' and $\times$ in `Labeling-Enh.'), leading to a slight performance degradation; a severe degradation is observed by adopting the non-enhanced $\f_m$ for both target sample selection and pseudo labels generation  (i.e., $\times$ in both `Selection-Enh.' and `Labeling-Enh.'), justifying the efficacy of implicit ensemble to provide better sampling indicator via improving the quality of predictive uncertainty to domain shift.

\begin{figure}
	\centering
	\subfigure[Feature discriminability] {
		\label{Fig:discriminability}
		\includegraphics[width=0.23\textwidth]{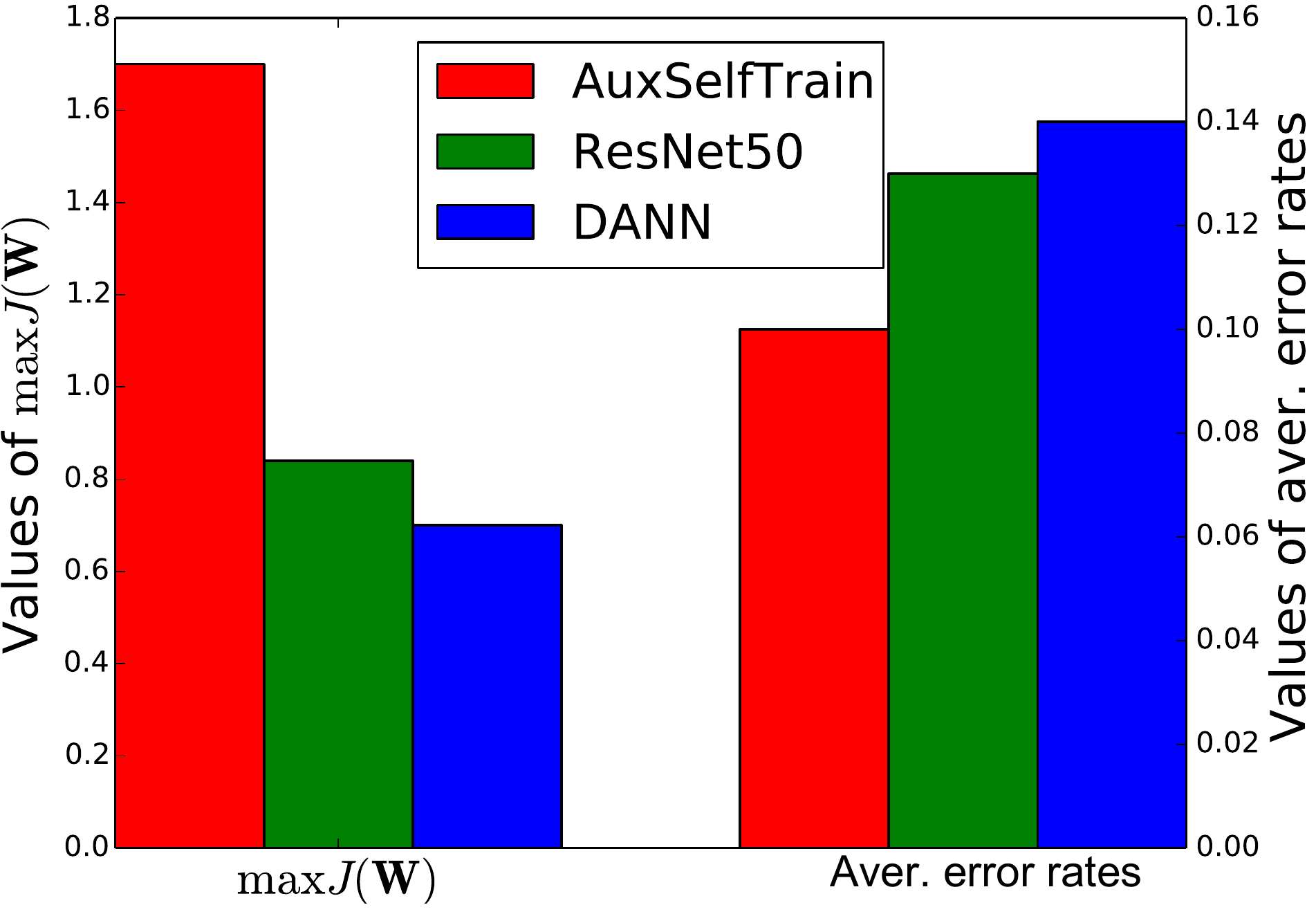}
	} \hfill    
	\subfigure[Results of various M] {
		\label{Fig:variousM}
		\includegraphics[width=0.215\textwidth]{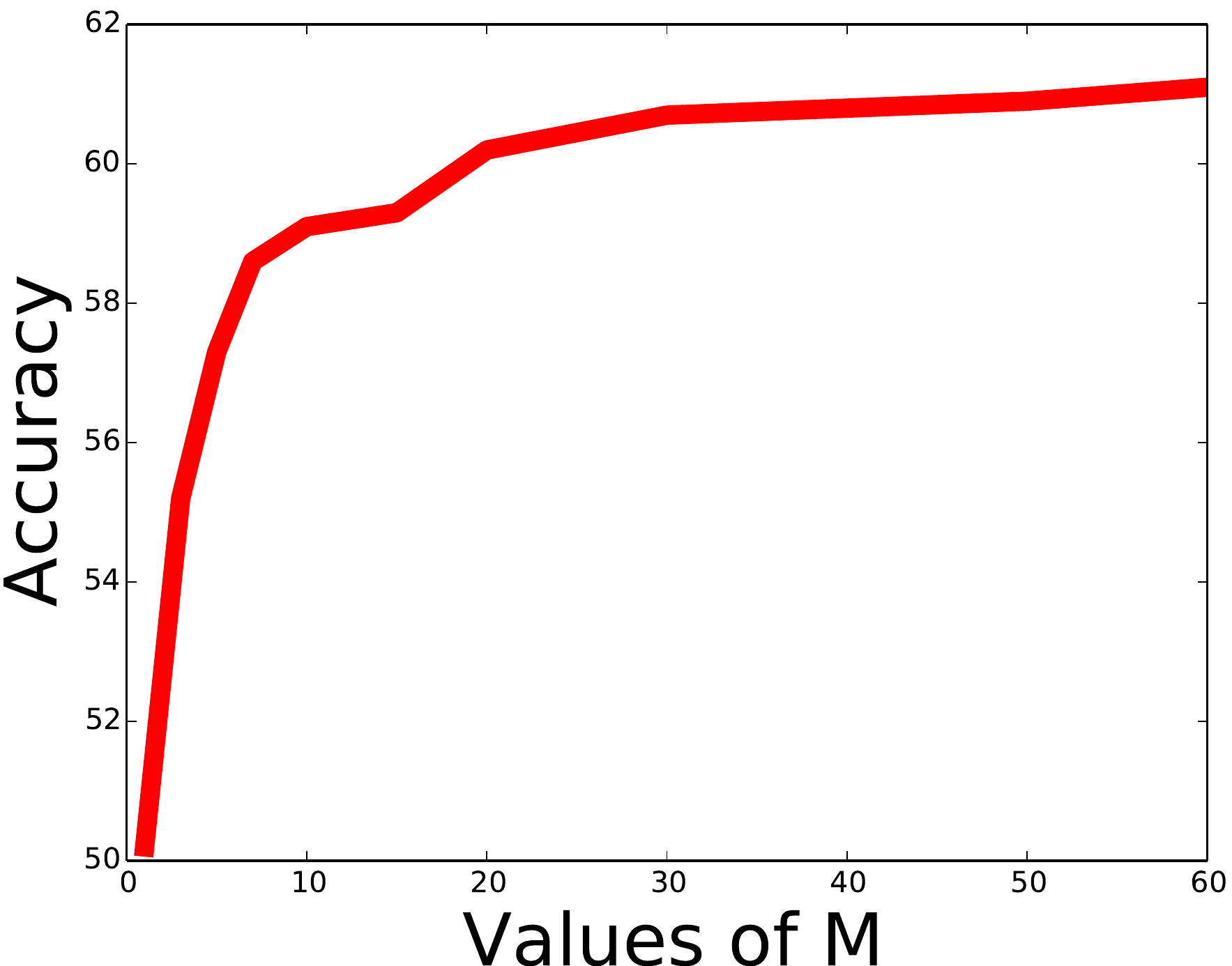} }
	\caption{(a) Two criteria measuring feature discriminability \cite{chen2019transferability}, where higher $\max J(\mathbf{W})$ and lower average error rates imply stronger discriminability. Definitions of the two criteria are given in the appendices. (b) Results of AuxSelfTrain with various $M$ on the Rw$\to$Cl task of OfficeHome dataset.}  
\end{figure}

\noindent \textbf{Analyses on number of intermediate domains}
We investigate the number of intermediate domains by setting $M$ as various values. 
As illustrated in Figure \ref{Fig:variousM}, the results get better as the $M$ increases in the beginning and gradually level off when $M$ reaches a relatively large value (e.g., $25$). The results are as expected since larger $M$ indicates smaller changes between consecutive domains. We empirically set $M$ as a large value, e.g., $50$, in experiments. 

\noindent  \textbf{Analyses on the advanced data augmentation}  Replacing the advanced data augmentation $\mathcal{A}$ \cite{cubuk2020randaugment}  in (\ref{Equ:overall_objective}) with a standard crop-and-flip one \cite{resnet} leads to degenerated results of 69.7\% and 77.5\% for the DA task on datasets of OfficeHome and VisDA-2017, respectively. To the best of our knowledge, we are the first one to investigate the advanced data augmentation \cite{cubuk2020randaugment} in DA problems. 

\noindent  \textbf{Analyses on target feature discriminability} Following \cite{chen2019transferability}, we investigate target feature discriminability on the VisDA-2017. As illustrated in Figure \ref{Fig:discriminability}, target feature discriminability in AuxSelfTrain is enhanced compared to the ResNet50 baseline; this distinguishes AuxSelfTrain from the seminal discrepancy minimization method \cite{dann}, where  target feature discriminability is deteriorated \cite{chen2019transferability}.



\subsubsection{Results}
As illustrated in Tables \ref{Table:visda}, \ref{Tab:home_close_da},  \ref{Tab:Office-Home-ssda3-vgg}, and \ref{Tab:domainnet_ssda}, our AuxSelfTrain achieve the new state of the art on all tested DA and SSDA benchmark datasets, justifying its efficacy on both DA and SSDA tasks and its generalization to different datasets and network structures.

\section{Discussions}
In contrast to popular distribution discrepancy minimization methods in DA \cite{dan,dann}, we propose gradual self-training of auxiliary models, which is empowered by evolving intermediate domains via a simple yet efficient sample selection strategy.
Together with \cite{kumar2020understanding}, we stress the importance of gradually evolving intermediate domains for the success of self-training on DA.
Two perspectives could be further investigated for the improvement. First, advanced indicators for sample selection could be explored. Moreover, the mix of source and target data could be conducted in other ways, such as pixel-level mixup \cite{zhang2017mixup}.  Additionally, although the correlation between domain shift and the maximum prediction probability has been empirically observed in \cite{ovadia2019can} and our paper, their underlying theoretical analysis is still an open problem and requires further investigation.

{\small
\bibliographystyle{ieee_fullname}
\bibliography{egbib}

\begin{thebibliography}{10}\itemsep=-1pt

\bibitem{ben2010theory}
Shai Ben-David, John Blitzer, Koby Crammer, Alex Kulesza, Fernando Pereira, and
  Jennifer~Wortman Vaughan.
\newblock A theory of learning from different domains.
\newblock {\em Machine learning}, 79(1-2):151--175, 2010.

\bibitem{ben2007analysis}
Shai Ben-David, John Blitzer, Koby Crammer, and Fernando Pereira.
\newblock Analysis of representations for domain adaptation.
\newblock In {\em Advances in neural information processing systems}, pages
  137--144, 2007.

\bibitem{bobu2018adapting}
Andreea Bobu, Eric Tzeng, Judy Hoffman, and Trevor Darrell.
\newblock Adapting to continuously shifting domains.
\newblock 2018.

\bibitem{caron2018deep}
Mathilde Caron, Piotr Bojanowski, Armand Joulin, and Matthijs Douze.
\newblock Deep clustering for unsupervised learning of visual features.
\newblock In {\em Proceedings of the European Conference on Computer Vision
  (ECCV)}, pages 132--149, 2018.

\bibitem{chen2019transferability}
Xinyang Chen, Sinan Wang, Mingsheng Long, and Jianmin Wang.
\newblock Transferability vs. discriminability: Batch spectral penalization for
  adversarial domain adaptation.
\newblock In {\em International Conference on Machine Learning}, pages
  1081--1090, 2019.

\bibitem{chen2019blending}
Ziliang Chen, Jingyu Zhuang, Xiaodan Liang, and Liang Lin.
\newblock Blending-target domain adaptation by adversarial meta-adaptation
  networks.
\newblock In {\em Proceedings of the IEEE Conference on Computer Vision and
  Pattern Recognition}, pages 2248--2257, 2019.

\bibitem{chopra2013dlid}
Sumit Chopra, Suhrid Balakrishnan, and Raghuraman Gopalan.
\newblock Dlid: Deep learning for domain adaptation by interpolating between
  domains.
\newblock In {\em ICML workshop on challenges in representation learning},
  volume~2, 2013.

\bibitem{cubuk2020randaugment}
Ekin~D Cubuk, Barret Zoph, Jonathon Shlens, and Quoc~V Le.
\newblock Randaugment: Practical automated data augmentation with a reduced
  search space.
\newblock In {\em Proceedings of the IEEE/CVF Conference on Computer Vision and
  Pattern Recognition Workshops}, pages 702--703, 2020.

\bibitem{cui2020hda}
Shuhao Cui, Xuan Jin, Shuhui Wang, Yuan He, and Qingming Huang.
\newblock Heuristic domain adaptation.
\newblock In {\em Advances in Neural Information Processing Systems}, 2020.

\bibitem{cui2020gradually}
Shuhao Cui, Shuhui Wang, Junbao Zhuo, Chi Su, Qingming Huang, and Qi Tian.
\newblock Gradually vanishing bridge for adversarial domain adaptation.
\newblock In {\em Proceedings of the IEEE/CVF Conference on Computer Vision and
  Pattern Recognition}, pages 12455--12464, 2020.

\bibitem{daume2010frustratingly}
Hal Daum{\'e}~III, Abhishek Kumar, and Avishek Saha.
\newblock Frustratingly easy semi-supervised domain adaptation.
\newblock In {\em Proceedings of the 2010 Workshop on Domain Adaptation for
  Natural Language Processing}, pages 53--59, 2010.

\bibitem{ding2018graph}
Zhengming Ding, Sheng Li, Ming Shao, and Yun Fu.
\newblock Graph adaptive knowledge transfer for unsupervised domain adaptation.
\newblock In {\em Proceedings of the European Conference on Computer Vision
  (ECCV)}, pages 37--52, 2018.

\bibitem{donahue2013semi}
Jeff Donahue, Judy Hoffman, Erik Rodner, Kate Saenko, and Trevor Darrell.
\newblock Semi-supervised domain adaptation with instance constraints.
\newblock In {\em Proceedings of the IEEE conference on computer vision and
  pattern recognition}, pages 668--675, 2013.

\bibitem{reverse_grad}
Yaroslav Ganin and Victor~S. Lempitsky.
\newblock Unsupervised domain adaptation by backpropagation.
\newblock In {\em Proceedings of the 32nd International Conference on Machine
  Learning, {ICML} 2015, Lille, France, 6-11 July 2015}, pages 1180--1189,
  2015.

\bibitem{dann}
Yaroslav Ganin, Evgeniya Ustinova, Hana Ajakan, Pascal Germain, Hugo
  Larochelle, Fran{\c{c}}ois Laviolette, Mario Marchand, and Victor Lempitsky.
\newblock Domain-adversarial training of neural networks.
\newblock {\em The Journal of Machine Learning Research}, 17(1):2096--2030,
  2016.

\bibitem{gong2019dlow}
Rui Gong, Wen Li, Yuhua Chen, and Luc~Van Gool.
\newblock Dlow: Domain flow for adaptation and generalization.
\newblock In {\em Proceedings of the IEEE Conference on Computer Vision and
  Pattern Recognition}, pages 2477--2486, 2019.

\bibitem{gan}
Ian Goodfellow, Jean Pouget-Abadie, Mehdi Mirza, Bing Xu, David Warde-Farley,
  Sherjil Ozair, Aaron Courville, and Yoshua Bengio.
\newblock Generative adversarial nets.
\newblock In {\em Advances in neural information processing systems}, pages
  2672--2680, 2014.

\bibitem{resnet}
Kaiming He, Xiangyu Zhang, Shaoqing Ren, and Jian Sun.
\newblock Deep residual learning for image recognition.
\newblock In {\em Proceedings of the IEEE conference on computer vision and
  pattern recognition}, pages 770--778, 2016.

\bibitem{hu2020unsupervised}
Lanqing Hu, Meina Kan, Shiguang Shan, and Xilin Chen.
\newblock Unsupervised domain adaptation with hierarchical gradient
  synchronization.
\newblock In {\em Proceedings of the IEEE/CVF Conference on Computer Vision and
  Pattern Recognition}, pages 4043--4052, 2020.

\bibitem{jiangbidirectional}
Pin Jiang, Aming Wu, Yahong Han, Yunfeng Shao, Meiyu Qi, and Bingshuai Li.
\newblock Bidirectional adversarial training for semi-supervised domain
  adaptation.
\newblock {\em IJCAI}, 2020.

\bibitem{can}
Guoliang Kang, Lu Jiang, Yi Yang, and Alexander~G Hauptmann.
\newblock Contrastive adaptation network for unsupervised domain adaptation.
\newblock In {\em Proceedings of the IEEE Conference on Computer Vision and
  Pattern Recognition}, pages 4893--4902, 2019.

\bibitem{kim2019self}
Seunghyeon Kim, Jaehoon Choi, Taekyung Kim, and Changick Kim.
\newblock Self-training and adversarial background regularization for
  unsupervised domain adaptive one-stage object detection.
\newblock In {\em Proceedings of the IEEE/CVF International Conference on
  Computer Vision}, pages 6092--6101, 2019.

\bibitem{kim2020attract}
Taekyung Kim and Changick Kim.
\newblock Attract, perturb, and explore: Learning a feature alignment network
  for semi-supervised domain adaptation.
\newblock In {\em European Conference on Computer Vision}, pages 591--607.
  Springer, 2020.

\bibitem{kumar2020understanding}
Ananya Kumar, Tengyu Ma, and Percy Liang.
\newblock Understanding self-training for gradual domain adaptation.
\newblock {\em ICML}, 2020.

\bibitem{lakshminarayanan2016simple}
Balaji Lakshminarayanan, Alexander Pritzel, and Charles Blundell.
\newblock Simple and scalable predictive uncertainty estimation using deep
  ensembles.
\newblock {\em arXiv preprint arXiv:1612.01474}, 2016.

\bibitem{mnist}
Yann LeCun, L{\'e}on Bottou, Yoshua Bengio, and Patrick Haffner.
\newblock Gradient-based learning applied to document recognition.
\newblock {\em Proceedings of the IEEE}, 86(11):2278--2324, 1998.

\bibitem{lee2013pseudo}
Dong-Hyun Lee.
\newblock Pseudo-label: The simple and efficient semi-supervised learning
  method for deep neural networks.
\newblock In {\em Workshop on challenges in representation learning, ICML},
  volume~3, page~2, 2013.

\bibitem{li2020online}
Da Li and Timothy Hospedales.
\newblock Online meta-learning for multi-source and semi-supervised domain
  adaptation.
\newblock {\em ECCV}, 2020.

\bibitem{dan}
Mingsheng Long, Yue Cao, Jianmin Wang, and Michael~I. Jordan.
\newblock Learning transferable features with deep adaptation networks.
\newblock In {\em Proceedings of the 32Nd International Conference on
  International Conference on Machine Learning - Volume 37}, ICML'15, pages
  97--105. JMLR.org, 2015.

\bibitem{cada}
Mingsheng Long, Zhangjie Cao, Jianmin Wang, and Michael~I Jordan.
\newblock Conditional adversarial domain adaptation.
\newblock In {\em Advances in Neural Information Processing Systems}, pages
  1640--1650, 2018.

\bibitem{mancini2018boosting}
Massimiliano Mancini, Lorenzo Porzi, Samuel Rota~Bul{\`o}, Barbara Caputo, and
  Elisa Ricci.
\newblock Boosting domain adaptation by discovering latent domains.
\newblock In {\em Proceedings of the IEEE Conference on Computer Vision and
  Pattern Recognition}, pages 3771--3780, 2018.

\bibitem{Mei2020instance}
Ke Mei, Chuang Zhu, Jiaqi Zou, and Shanghang Zhang.
\newblock Instance adaptive self-training for unsupervised domain adaptation.
\newblock In {\em ECCV}, pages 415--430. Springer International Publishing,
  2020.

\bibitem{nguyen2015deep}
Anh Nguyen, Jason Yosinski, and Jeff Clune.
\newblock Deep neural networks are easily fooled: High confidence predictions
  for unrecognizable images.
\newblock In {\em Proceedings of the IEEE conference on computer vision and
  pattern recognition}, pages 427--436, 2015.

\bibitem{ovadia2019can}
Yaniv Ovadia, Emily Fertig, Jie Ren, Zachary Nado, David Sculley, Sebastian
  Nowozin, Joshua~V Dillon, Balaji Lakshminarayanan, and Jasper Snoek.
\newblock Can you trust your model's uncertainty? evaluating predictive
  uncertainty under dataset shift.
\newblock {\em NIPS}, 2019.

\bibitem{transfer_survey}
Sinno~Jialin Pan, Qiang Yang, et~al.
\newblock A survey on transfer learning.
\newblock {\em IEEE Transactions on knowledge and data engineering},
  22(10):1345--1359, 2010.

\bibitem{peng2019moment}
Xingchao Peng, Qinxun Bai, Xide Xia, Zijun Huang, Kate Saenko, and Bo Wang.
\newblock Moment matching for multi-source domain adaptation.
\newblock In {\em Proceedings of the IEEE International Conference on Computer
  Vision}, pages 1406--1415, 2019.

\bibitem{peng2017visda}
Xingchao Peng, Ben Usman, Neela Kaushik, Judy Hoffman, Dequan Wang, and Kate
  Saenko.
\newblock Visda: The visual domain adaptation challenge.
\newblock {\em arXiv preprint arXiv:1710.06924}, 2017.

\bibitem{sagi2018ensembles}
Omer Sagi and Lior Rokach.
\newblock Ensemble learning: A survey.
\newblock {\em Wiley Interdisciplinary Reviews: Data Mining and Knowledge
  Discovery}, 8(4):e1249, 2018.

\bibitem{ssda_mme}
Kuniaki Saito, Donghyun Kim, Stan Sclaroff, Trevor Darrell, and Kate Saenko.
\newblock Semi-supervised domain adaptation via minimax entropy.
\newblock In {\em 2019 {IEEE/CVF} International Conference on Computer Vision,
  {ICCV} 2019, Seoul, Korea (South), October 27 - November 2, 2019}, pages
  8049--8057. {IEEE}, 2019.

\bibitem{mcd}
Kuniaki Saito, Kohei Watanabe, Yoshitaka Ushiku, and Tatsuya Harada.
\newblock Maximum classifier discrepancy for unsupervised domain adaptation.
\newblock In {\em Proceedings of the IEEE Conference on Computer Vision and
  Pattern Recognition}, pages 3723--3732, 2018.

\bibitem{vgg}
Karen Simonyan and Andrew Zisserman.
\newblock Very deep convolutional networks for large-scale image recognition.
\newblock {\em arXiv preprint arXiv:1409.1556}, 2014.

\bibitem{snell2017prototypical}
Jake Snell, Kevin Swersky, and Richard Zemel.
\newblock Prototypical networks for few-shot learning.
\newblock In {\em Advances in Neural Information Processing Systems}, pages
  4077--4087, 2017.

\bibitem{sohn2020fixmatch}
Kihyuk Sohn, David Berthelot, Chun-Liang Li, Zizhao Zhang, Nicholas Carlini,
  Ekin~D Cubuk, Alex Kurakin, Han Zhang, and Colin Raffel.
\newblock Fixmatch: Simplifying semi-supervised learning with consistency and
  confidence.
\newblock {\em arXiv preprint arXiv:2001.07685}, 2020.

\bibitem{deep_coral}
Baochen Sun and Kate Saenko.
\newblock Deep coral: Correlation alignment for deep domain adaptation.
\newblock In {\em European Conference on Computer Vision}, pages 443--450.
  Springer, 2016.

\bibitem{srdc}
Hui Tang, Ke Chen, and Kui Jia.
\newblock Unsupervised domain adaptation via structurally regularized deep
  clustering.
\newblock In {\em Proceedings of the IEEE/CVF Conference on Computer Vision and
  Pattern Recognition}, pages 8725--8735, 2020.

\bibitem{home}
Hemanth Venkateswara, Jose Eusebio, Shayok Chakraborty, and Sethuraman
  Panchanathan.
\newblock Deep hashing network for unsupervised domain adaptation.
\newblock In {\em Proceedings of the IEEE Conference on Computer Vision and
  Pattern Recognition}, pages 5018--5027, 2017.

\bibitem{xie2015transfer}
Michael Xie, Neal Jean, Marshall Burke, David Lobell, and Stefano Ermon.
\newblock Transfer learning from deep features for remote sensing and poverty
  mapping.
\newblock {\em arXiv preprint arXiv:1510.00098}, 2015.

\bibitem{xie2020unsupervised}
Qizhe Xie, Zihang Dai, Eduard Hovy, Thang Luong, and Quoc Le.
\newblock Unsupervised data augmentation for consistency training.
\newblock {\em Advances in Neural Information Processing Systems}, 33, 2020.

\bibitem{xu2020reliable}
Renjun Xu, Pelen Liu, Liyan Wang, Chao Chen, and Jindong Wang.
\newblock Reliable weighted optimal transport for unsupervised domain
  adaptation.
\newblock In {\em Proceedings of the IEEE/CVF Conference on Computer Vision and
  Pattern Recognition}, pages 4394--4403, 2020.

\bibitem{yao2015semi}
Ting Yao, Yingwei Pan, Chong-Wah Ngo, Houqiang Li, and Tao Mei.
\newblock Semi-supervised domain adaptation with subspace learning for visual
  recognition.
\newblock In {\em Proceedings of the IEEE conference on Computer Vision and
  Pattern Recognition}, pages 2142--2150, 2015.

\bibitem{zhang2017mixup}
Hongyi Zhang, Moustapha Cisse, Yann~N Dauphin, and David Lopez-Paz.
\newblock mixup: Beyond empirical risk minimization.
\newblock {\em ICLR}, 2018.

\bibitem{zhang2020label}
Yabin Zhang, Bin Deng, Kui Jia, and Lei Zhang.
\newblock Label propagation with augmented anchors: A simple semi-supervised
  learning baseline for unsupervised domain adaptation.
\newblock In {\em European Conference on Computer Vision}, pages 781--797.
  Springer, 2020.

\bibitem{zhang2019bridging}
Yuchen Zhang, Tianle Liu, Mingsheng Long, and Michael Jordan.
\newblock Bridging theory and algorithm for domain adaptation.
\newblock In {\em International Conference on Machine Learning}, pages
  7404--7413, 2019.

\bibitem{symnets}
Yabin Zhang, Hui Tang, Kui Jia, and Mingkui Tan.
\newblock Domain-symmetric networks for adversarial domain adaptation.
\newblock In {\em Proceedings of the IEEE Conference on Computer Vision and
  Pattern Recognition}, pages 5031--5040, 2019.

\bibitem{zhao2019learning}
Han Zhao, Remi Tachet~des Combes, Kun Zhang, and Geoffrey~J Gordon.
\newblock On learning invariant representation for domain adaptation.
\newblock {\em arXiv preprint arXiv:1901.09453}, 2019.

\bibitem{zhou2004learning}
Dengyong Zhou, Olivier Bousquet, Thomas~N Lal, Jason Weston, and Bernhard
  Sch{\"o}lkopf.
\newblock Learning with local and global consistency.
\newblock In {\em Advances in neural information processing systems}, pages
  321--328, 2004.

\bibitem{zou2018unsupervised}
Yang Zou, Zhiding Yu, BVK Vijaya~Kumar, and Jinsong Wang.
\newblock Unsupervised domain adaptation for semantic segmentation via
  class-balanced self-training.
\newblock In {\em Proceedings of the European conference on computer vision
  (ECCV)}, pages 289--305, 2018.

\end{thebibliography}
}

\end{document}